\documentclass[11pt]{article}

\usepackage[final]{acl}

\usepackage[utf8]{inputenc}

\usepackage{times}
\usepackage{latexsym}
\usepackage{booktabs}
\usepackage{xurl}
\usepackage{tcolorbox}
\tcbuselibrary{breakable}
\usepackage{csquotes}
\usepackage{graphicx}
\usepackage{longtable}
\usepackage{makecell}
\usepackage{subcaption}
\usepackage{arydshln}
\usepackage{amsmath}
\usepackage{multirow}
\usepackage{float}
\usepackage{tabularx,ragged2e,adjustbox}
\usepackage{todonotes}

\usepackage{pifont}
\newcommand{\cmark}{\textcolor{green!60!black}{\ding{51}}} 
\newcommand{\xmark}{\textcolor{red}{\ding{55}}}            

\usepackage{listings}
\usepackage{xcolor}

\lstdefinestyle{promptstyle}{
    basicstyle=\ttfamily\scriptsize,
    breaklines=true,
    breakatwhitespace=false,
    columns=fullflexible,
    keepspaces=true,
    frame=single,
    framesep=3pt,
    xleftmargin=3pt,
    xrightmargin=3pt,
    backgroundcolor=\color{gray!5},
    showstringspaces=false,
}

\usepackage[T1]{fontenc}

\usepackage{microtype}

\usepackage{inconsolata}
\usepackage{amssymb}
\usepackage{fontawesome5}

\usepackage{graphicx}

\title{What Is The Political Content in LLMs' Pre- and Post-Training Data?}

\author{First Author \\
  Affiliation / Address line 1 \\
  Affiliation / Address line 2 \\
  Affiliation / Address line 3 \\
  \texttt{email@domain} \\\And
  Second Author \\
  Affiliation / Address line 1 \\
  Affiliation / Address line 2 \\
  Affiliation / Address line 3 \\
  \texttt{email@domain} \\}

\author{
 \textbf{Tanise Ceron\textsuperscript{1}},
 \textbf{Dmitry Nikolaev\textsuperscript{2}},
  \textbf{Dominik Stammbach\textsuperscript{3}},
  \textbf{Debora Nozza\textsuperscript{1}},
\\
\\
 \textsuperscript{1}Bocconi University,
 \textsuperscript{2} University of Manchester,
 \textsuperscript{3} Princeton University,
\\
 \small
   \textbf{Correspondence:} \href{mailto:tanise.ceron@unibocconi.it}{tanise.ceron@unibocconi.it}
   \\
   \href{https://huggingface.co/datasets/tceron/stancepol}{\raisebox{-0.15em}{\includegraphics[height=1em]{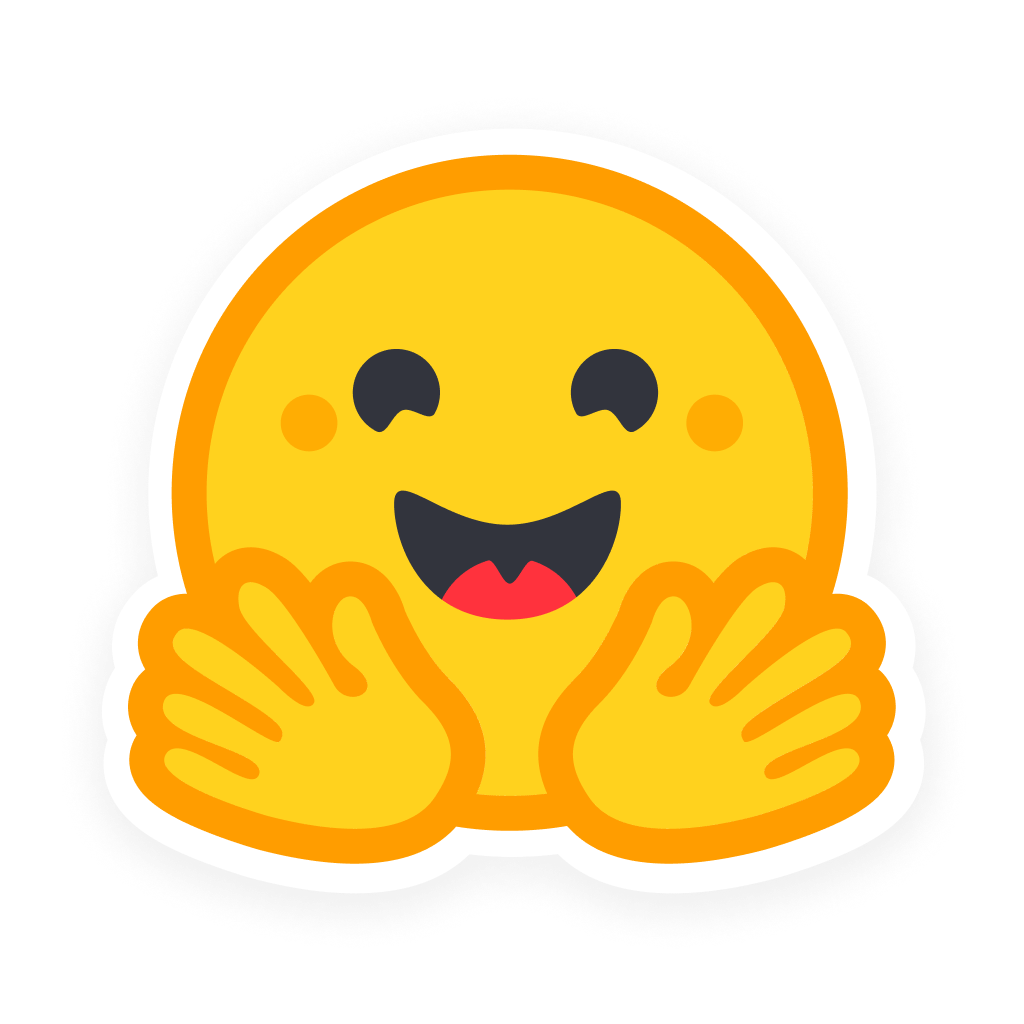}}~Data} \quad
   \href{https://github.com/tceron/political_content_training_data}{\faGithub~Code}
}

\begin{document}
\maketitle
\begin{abstract}

Large language models (LLMs) reflect politically-slanted opinions in their generated text. Even though it is widely assumed that model behavior stem from training data, there has been no study quantifying the extent to which political content is part of the training data.
To bridge this gap, we aim to directly estimate (1)~the proportion of politically engaged texts in training data, (2)~respective data imbalance, (3)~cross-dataset similarity, and (4)~correlations between data composition and model behaviour.
We analyze the political content of pre- and post-training datasets of open-source LLMs, combining large-scale sampling, political-leaning classification, and stance detection. We find that all LLM training datasets are systematically skewed towards left-leaning content, with pre-training containing more politically engaged than post-training corpora. We further observe a strong correlation between political stances in training data and model behavior, which is present already in most base models and persists across post-training stages.
These findings highlight the role of data composition in correlating with model behavior and motivate the need for greater data transparency as a means to understand and monitor model behavior. 

\end{abstract}

\section{Introduction}

\label{sec:intro}

LLM-powered chatbots have gained widespread adoption, 
with more and more people using them as sources of information, commentary, and practical advice \citep{ouyang-etal-2023-shifted,bassignana-etal-2025-ai}. 
In particular, \citet{luettgau2025political} observe not only that around 13\% of eligible UK voters may have used conversational AI to inform their electoral choices, but also that such use enhanced political knowledge to a degree comparable with web searches. Beyond knowledge acquisition, \citet{potter-etal-2024-hidden} and \citet{salvi2025conversational} show that LLMs can influence users’ decision-making and political beliefs by acting as information intermediaries. This reliance on LLMs for information gathering raises concerns about whether these systems present balanced, objective, and fair perspectives \citep{balkin2017digital,10.1145/3531146.3533088}. Prior work has shown that LLMs exhibit systematic political leanings, often aligning with left-leaning or libertarian viewpoints \citep{motoki2023more, hartmann2023political, rottger-etal-2024-political, ceron-etal-2024-beyond, nikolaev2026llmsreadyhardchoices}. However, how these opinions are formed within language models remains underexplored. 

Although researchers largely agree
that a deep understanding of training data is essential for interpreting machine learning models \citep{10.1145/3287560.3287596,10.1145/3442188.3445918,kaddour2023challenges}, most existing research on political opinions in LLMs focuses on evaluating model outputs or alignment techniques \citep{stammbach-etal-2024-aligning,weeber-etal-2026-political,kimlinear,agiza2024politune}. 
This perspective alone cannot identify or quantify sources of political opinions, making it difficult to understand the behavior of models or prevent information manipulation at scale \citep{waight2026state,graf2026pretraining}.  
We address this gap by examining political content in the underlying training data -- and its relationship to model behavior. We analyze open-source models and their associated pre- and post-training datasets, focusing on the political leaning of the content, statistical distribution of different stances, and their correlation with model attitudes across eight major policy issues.

Building on prior findings, we formulate the following research questions to guide our analysis:
\begin{itemize}
    \itemsep-0.25em 
    \item[\textbf{R1}] Models reflect left/progressive-leaning views \citep{hartmann2023political,motoki2023more,ceron-etal-2024-beyond,nikolaev2026llmsreadyhardchoices}. How high is the proportion of left-leaning documents in the training data? 
    \item[\textbf{R2}] Training datasets differ substantially in their composition \citep{mansour2025measuring}. Is there variance in the political content present in different pre-training corpora? 
    \item[{\textbf{R3}}] Gender and racial biases in models have a strong correlation with the training data \citep{garg2018word,leavy2020mitigating}. How strongly does that hold for political stances which are not only at the surface level of language (e.g. pronouns and name associations), but are entire discursive constructs? 
    \item[{\textbf{R4}}] Pre-training dominates the learning process, with post-training primarily refining model representations \citep{zhou2023lima,qi2025safety}~-- unless alignment is intentional. How do base models differ from post-trained models and their training datasets? 
\end{itemize}
\vspace{-10pt}
\noindent We test these questions through a combination of large-scale sampling, political leaning classification, and stance analysis of both training data and model outputs in Pythia-12B \citep{biderman2023pythia}, Falcon-11B \cite{malartic2024falcon2}, and different checkpoints of OLMO3-7B \citep{olmo3-2025}. 


Overall, our results show that left-leaning documents are 1.66 to 6.29 times as prevalent as right-leaning documents across all training datasets. Moreover, pre-training corpora contain 1.7 to 2.7 times as much politically aligned content as post-training datasets. Furthermore, different pre-training datasets exhibit highly similar stances towards 8 policy issues (e.g. social welfare state, migration, etc), suggesting that variations in data curation may have only limited impact on aggregate political properties. Finally, we find a very strong correlation ($r=0.88$, $p<0.05$) between the political stances present in training data and those expressed by models across those 8 policy issues.

\paragraph{Contributions.} We present the first general data-centric study directly linking training data composition to political behavior in LLMs.  Methodologically, we contribute with (i) a validated ensemble classifier for document-level political leaning and a stance classifier validated against a newly annotated dataset with training data documents (\textsc{stancepol}), both applied at corpus scale across pre- and post-training data; and (ii) we examine three model families with non-overlapping training corpora, allowing us to rule out that the observed skew is only artifact of shared LLM training datasets. Empirically, we quantify the proportion of politically engaged documents across training datasets in terms of political ideology and political stance towards 8 policy issues. We further show this skew correlates strongly with most base-model -- not just aligned-model -- stance. Together, these results motivate data-transparency enforced at the policy level given that training data can be served as a lever for monitoring LLMs' behavior \footnote{Code and data are available in the links at the header.}.

\begin{table*}[h!]
\centering
\resizebox{\textwidth}{!}{%
\begin{tabular}{cccc c ccc}
\cmidrule(lr){1-4} \cmidrule(lr){6-8}
\multicolumn{4}{c}{{\textbf{Documents for analysis}}} & & \multicolumn{3}{c}{{\textbf{Classification results}}} \\
\cmidrule(lr){1-4} \cmidrule(lr){6-8}
\textbf{Dataset name} & \textbf{Total size} & \textbf{Training and model} & \textbf{\# Docs} & & \textbf{\# (\%) Left} & \textbf{\# (\%) Right} & \textbf{Ratio L/R [CI]} \\
\cmidrule(lr){1-4} \cmidrule(lr){6-8}
\makecell{ \textsc{\textbf{Dolma3}} \\ \textsc{dolma3-mix-6T-1025-7B}} & 4.05TB & \makecell{Pre-training \\ \textsc{Olmo-3-1025-7B}}  & 473,837 & & 18,196 (3.84\%) & 10,260 (2.17\%) & 1.77 [1.73, 1.82] \\
\makecell{\textsc{\textbf{Dolmino3}} \\ \textsc{dolma3-dolmino-mix-100B-1025}} & 180GB & \makecell{Mid-pre-training \\ \textsc{Olmo-3-1025-7B}} & 188,107 & & 5,758 (3.06\%) & 3,472 (1.85\%) & 1.66	[1.59, 1.73] \\
\makecell{ \textsc{\textbf{RefinedWeb}} \\ \textsc{falcon-refinedweb}} & 1.68TB & \makecell{Pre-training \\ \textsc{falcon2-11B}}  & 194,335 & & 6,089 (3.13\%) & 2,676 (1.38\%) & 2.27	[2.17, 2.38] \\
\makecell{ \textsc{\textbf{The Pile}} \\ \textsc{the\_pile\_deduplicated}} & 451GB & \makecell{Pre-training \\ \textsc{pythia-12b-deduped}} & 174,849 & & 7,807 (4.46\%) & 3,060 (1.75\%) & 2.55 [2.45, 2.66] \\
\makecell{\textsc{\textbf{SFT-Mix}}  \\ \textsc{Dolci-Instruct-SFT} } & 3GB & \makecell{Supervised finetuning (SFT) \\ \textsc{Olmo-3-7B-SFT}} & 144,927 & & 3,289 (2.27\%) & 523 (0.36\%) & 6.28 [5.75, 6.90] \\
\makecell{\textsc{\textbf{DPO-Mix}} \\ \textsc{Dolci-Instruct-DPO} } & 810MB & \makecell{Direct Preference Optim. (DPO) \\ \textsc{Olmo-3-7B-DPO}} & 223,736 & & 4,205 (1.88\%) & 941 (0.42\%) & 4.46 [4.17, 4.80] \\
\cmidrule(lr){1-4} \cmidrule(lr){6-8}
\end{tabular}
}
\caption{Training datasets from \textsc{Olmo3}, \textsc{Pythia}, and \textsc{Falcon2}, with the number of sampled documents and their proportion classified as left- or right-leaning. The remaining documents in the classification results were classified as \textit{neutral}. Ratio L/R denotes the ratio of left-leaning to right-leaning document counts, with 95\% bootstrap confidence intervals in brackets (10,000 resamples of the documents in each corpus).}
\label{tab:olmo-datasets}
\end{table*}

\section{Data}
\label{sec:data-training-data}

\subsection{Target datasets}

We examine three models, and summarize the corresponding datasets used to train these models in Table \ref{tab:olmo-datasets}. \textsc{Dolma3} and \textsc{Dolmino3} are the datasets used in the pre-training and mid-pre-training stages of \textsc{Olmo3-Base}, respectively. As detailed in the report \citep{olmo3-2025}, the stages differ in the allocation of floating-point operations (FLOPs) and the data: pre-training relies on a broader, less filtered collection of web-scraped documents (\textsc{Dolma}) and uses 90\% of training FLOPs, whereas mid-pre-training continues next-token prediction on a higher-quality, filtered dataset (\textsc{Dolmino}) for the remaining 10\% of training FLOPs. \textsc{RefinedWeb} is used to pretrain \textsc{Falcon2-11B}, whereas the deduplicated version of \textsc{ThePile} is used to train \textsc{Pythia-12B-dedup}.

For our investigations, we select models that have open and maximally non-overlapping 
training data.
Table~\ref{tab:final-dataset-comparison} in the Appendix reports data overlaps across models. E.g., \textsc{Olmo3} \& \textsc{SmolLM3} \citep{bakouch2025smollm3} 
and \textsc{Olmo3} \& \textsc{Marin-8B} overlap in seven datasets, while \textsc{Olmo3} \& \textsc{Apertus-8B} \citep{hernandez2025apertus}
overlap in two large datasets. We, therefore, select three models that contain mostly English data and whose training data do not overlap: \textsc{Olmo3} \citep{olmo3-2025}, \textsc{Falcon2} \cite{malartic2024falcon2}, and \textsc{Pythia} \citep{biderman2023pythia}.

Among these, we only have access to intermediate checkpoints and post-training datasets for the \textsc{Olmo3} model.\footnote{\textsc{Falcon2-11B} has an instruct-fine-tuned version trained with vision examples, and \textsc{Pythia} does not have any instruct-fine-tuned model.}
\textsc{SFT3} is the dataset used for supervised fine-tuning, the first post-training stage of the base model, where it learns to solve a variety of traditional NLP tasks expressed in the dialogue form. This stage also relies on the standard next-token prediction loss. 
\textsc{DPO3} is the dataset used in the second post-training phase, where the model learns human preferences with DPO \citep{rafailov2023direct}, a contrastive preference-optimization method.\footnote{We do not analyze the data used in the subsequent post-training phase of the Instruct version, as it mainly consists of mathematical or coding content, see \url{allenai/Dolci-Instruct-RL}.}

\subsection{Sampling}

Since investigating the content of the entire pretraining datasets is not feasible due to computational resource constraints, we resort to sampling to retrieve a representative set of documents for our analysis of the pre-training data. For that, we first fix a sample size of $N$ documents and then use the classic version of the reservoir 
sampling algorithm \citep{vitter1985reservoir}, which extracts uniform random samples from the base dataset in an online fashion. For the purposes of sampling, we exclude the subsets that focus on code and mathematical problems, based on the model card description.

$N$ is set to 500k for \textsc{Dolma3} and 200k for the remaining datasets, except for \textsc{DPO3}, which only has approximately 255k documents, considerably fewer than the other datasets, and which we process in its entirety.
We further remove non-English documents with a fastText-based language classifier (threshold$>$0.7$=$English). We present statistics for each dataset in Table~\ref{tab:olmo-datasets}.

\subsection{Validation datasets}

Below we describe the datasets used to evaluate the classifiers used in this study. 

\paragraph{\textsc{newslean}.} To evaluate our classifier political leaning classifier, we use the dataset introduced by \citet{braghieri2024level}, comprising 4,434 articles drawn from the 94 most-visited US news outlets in 2019 (see Appendix~\ref{appendix:news-article-dataset} for details). This dataset provides article-level human annotations of political leaning, in contrast to existing alternatives \citep{baly2020we,liu2021transformer}, which infer document-level labels from source-level heuristics. 

\begin{table}[]
    \centering
    \small
    \begin{tabular}{lr}
    \toprule
    policy & Cohen's $k$ \\
    \midrule
    expanded environ protection & 0.887 \\
    restrictive migration policy & 0.801 \\
    liberal economic policy & 0.696 \\
    law and order & 0.658 \\
    expanded social welfare & 0.623 \\
    liberal society & 0.552 \\
    open foreign policy & 0.539 \\
    restrictive financial policy & 0.483 \\
    Average & 0.655 \\
    \bottomrule
    \end{tabular}
    \caption{Cohen's kappa between 2 annotators in the task of annotating the stance of documents towards policy domains.}
    \label{tab:annotations-stance}
\end{table}

\paragraph{\textsc{stancepol}.} We construct a new dataset with 200 documents sampled directly from the training dataset \textsc{Dolma}. We annotate the documents for policy issues with support/reject/neutral labels. Two annotators annotate the documents following guidelines found in Appendix \ref{appendix:stance-annotations}). As Table \ref{tab:annotations-stance} shows, Cohen’s $\kappa$ between annotators and policy issues falls within the range of moderate agreement ($\kappa$ = 0.65), which is considered good for a subjective task such as political stance detection \citep{wong2021cross}. The highest agreement is in \textit{Expanded Environment Protection} ($\kappa$ = 0.88) and the lowest in \textit{Restrictive Financial Policy} ($\kappa$ = 0.48). Finally, we add a new layer of labels in \textsc{stancepol}. If a document is, for example, in favor of liberal society, it is considered a "left" leaning document while if it is against liberal society, then it is considered "right" leaning. The conversion label is found in Appendix Table \ref{tab:directions}. \textsc{stancepol} for political leaning contains 81 left- documents and 64 right-leaning documents. 

\section{Methods}
\label{sec:political-content}

In this section, we present the methods we applied for identifying political leaning present in documents from the training datasets (\S~\ref{subsec:document-classification}) and for analyzing topical content of left- and right-leaning documents (\S~\ref{subsec:topic-analysis}). 

\subsection{Political leaning classification}
\label{subsec:document-classification}

Our first step involves classifying the target documents into left-leaning, right-leaning, and neutral. We acknowledge that the left–right axis represents a simplification of the broader ideological space. However, this abstraction is well-established in political science and continues to serve as a standard methodological tool for analyzing political text and speech, especially at scale \citep{jolly2022chapel,lehmann2022manifesto,lauderdale2016measuring, vafa-etal-2020-text, Goet_2019}.
 
We adopt a three-way classification of documents as left-leaning loaded, right-leaning loaded, and neutral. We classify documents as neutral if they either lack political content altogether due to their topic or do not express an evident right- or left-leaning stance.

We acknowledge that this is a simplification too, as (i)~a centrist political position is an ideology in its own right, and (ii)~politically \enquote{neutral} texts are at times not completely neutral. The distinction between a purely apolitical document, however, and one that has some political content but is ideologically neutral, is hard to make consistently even for human experts, e.g., a text describing the system of government or taxation of a particular country may be perceived as purely informational and therefore apolitical; however, only a very slight change is needed to change it into a text focusing on \textit{inefficiencies} in this system, which will make it a politically-engaged, and arguably ideological, document. 
Ultimately, we are interested in left- and right-leaning stances around policy issues. Hence, we use a simplified 3-label classification scheme and discard both non-political and centrist texts. For additional validation, we conduct a topic-modeling analysis of \enquote{neutral} documents, which shows that they are largely non-political in content
(see Appendix~\ref{appendix:neutral-documents}).

\begin{table}[t]
\centering
\small
\begin{tabular}{llrrrr}
\toprule
 & & prec. & rec. & f1 & support \\
\midrule
\multirow{3}{*}{\textsc{NewsLean}}
 & left & 0.83 & 0.59 & 0.69 & 1148 \\
 & neutral & 0.65 & 0.86 & 0.74 & 1148 \\
 & right & 0.80 & 0.75 & 0.77 & 627 \\
\midrule
\multirow{2}{*}{\textsc{StancePol}}
 & left & 0.97 & 0.81 & 0.88 & 81 \\
 & right & 0.85 & 0.89 & 0.87 & 64 \\
\bottomrule
\end{tabular}
\caption{Results of the left-neutral-right classification with the best prompt and the ensemble setup with \textsc{Qwen3.5-27B} and \textsc{Gemma4-31B} on the \textsc{NewsLean} dataset, and validation of the ensemble classifier on the \textsc{StancePol} dataset.}
\label{tab:classification-results-ensemble}
\end{table}

\paragraph{Classification pipeline.} 

We rely on mid-size LLMs to classify documents at scale. After extensive evaluation of different models and prompt versions on a sample of 2923 documents from \textsc{newslean}, with a focus on prioritizing recall over precision (since false positives have another chance of being filtered out at the stage of stance analysis), we settled on an ensemble combination of \textsc{Gemma4-31B} and \textsc{Qwen3.5-27B}, with the Gemma model used a tie-breaker. This combination reached an F1-macro score of 0.73 with an average recall across the \enquote{left} and \enquote{right} categories of 0.67 (cf.\ Table~\ref{tab:prompt-templates-results} in Appendix).\footnote{The literature on political ideology classification has largely utilized fine-tuned encoder-only models, such as those adopted by \citet{nikolaev-etal-2023-multilingual}. However, neither they nor fine-tuned ModernBERT \citep{warner-etal-2025-smarter} performed well on documents from our pool.} 

We run the ensemble classifier with the best prompt on \textsc{stancepol} to validate our classifier on blogs and web documents besides news articles. As shown in Table \ref{tab:classification-results-ensemble}, results are higher than in the \textsc{newslean} dataset with average recall between left and right reaching 0.85. Overall, the recall of the ensemble classifier is higher in right-leaning documents in both datasets, where it achieves 0.75 vs.\ 0.59 in the \textsc{newslean} and 0.89 vs.\ 0.81 in the \textsc{stancepol}. See detailed results in Appendix~\ref{appendix:left-right-classification}.

\subsection{Topic modeling}
\label{subsec:topic-analysis}

We further run topic modeling to investigate the extent to which left- and right-leaning texts engage with the same issues. We extract topics using the BERTopic framework \cite{grootendorst2022bertopic}, randomly sampling 3000 left- and right-leaning documents from each dataset, embedding the first 32k tokens of each document using Qwen3 Embedding 4B \citep{qwen3embedding}, and splitting the documents pooled across datasets into 60 clusters using CLARA \citep{park2009simple}\footnote{As implemented in the \texttt{sklearn\_extra} package.}. We then extract 20 most frequent topics per dataset and political leaning and inspect their overlap.

\subsection{Political stances in LLMs}
\label{subsec:model-stance}

We compute political stances in models to then correlate them with stances present in the training data. This analysis allows us to identify where such stances are formed. We deliberately choose the term stances rather bias because the former refers to the presence of consistent detectable opinions in the generated text whereas the latter is often perceived as something that needs to be corrected \citep{mehrabi2021survey}. We define stances in LLMs as the consistency in supporting or opposing a certain topic or policy issue. 

We largely follow the method of \citet{ceron-etal-2024-beyond}: a dataset containing 239 statements compiled from voting advice applications from seven European countries. Each statement is annotated as to whether it belongs to one or more policy issues within the eight broad policy vectors which form the target of the analysis (expanded environment protection, liberal society, liberal economic policy, open foreign policy, expanded social welfare, law and order, restrictive financial policy, and restrictive migration policy). For example, agreeing or disagreeing with the statement \enquote{Childcare being free for all parents for at least three days a week} means being in favor or against the general position of \textit{Expanded social welfare state}. 
 
The dataset contains different statement formulations (3 paraphrases, 1 negation and 1 semantically inverted version of the original statement). We run all 6 versions of the statements, combined with the 12 prompt instructions found in \cite{ceron-etal-2024-beyond}, resulting in 72 versions of the same statement. We furthermore sample the generated answer on the same prompt 30 times with the temperature set to 1, thus reaching 2,160 answers for each statement.

We compute the stance using:
\begin{equation}
Stance_{(p,m)}
= \frac{1}{\lvert \mathcal{J}p\rvert}\sum{j \in \mathcal{J}p} a{p,j}, r_{m,j}
\end{equation}
\noindent We denote by $Stance_{(p,m)}\in{-1,1}$ the attitude toward policy $p$ encoded in statement $j$; $r_{m,j}\in[-1,1]$ is model $m$'s response, and $\mathcal{J}_p$ is the set of items with nonzero annotations for policy $p$ that model $m$ answered. When the stance is close to 1 or $-$1, this means that the model is very consistent in the stance towards the statement. If the stance is close to 0, the model oscillates between agreeing and disagreeing with the statement.

\subsection{Political stances in training data}

To obtain comparable estimates of how strongly models' outputs are correlated with training data alone, we further validate a document stance classifier. 

\paragraph{Classification pipeline.} Using \textsc{stancepol}, we evaluate accuracy of \textsc{GPT-5.4mini} and \textsc{Qwen3.6-27B} using the ground truth labels in \textsc{stancepol}. Both reach an f1-macro score of 0.81 (see Appendix~\ref{appendix:stance-classification}). 
The average precision and recall across policy domains is 77\% and 88\% respectively. Precision across issues ranges from 85\% for \textit{Expanded Environment Protection} to 69\% for \textit{Law and Order} (see Appendix Table~\ref{tab:classification-stance}). We use 
\textsc{GPT-5.4mini} as our stance classifier and extract stances of documents towards the 8 policy issues. We use these labels to compute the general stance of the training data: $S_{pol} = (S - O)/(S + O)$, where $S$ is the total number of documents supporting and $O$ is the total number of documents opposing the policy vector.

\begin{figure*}
    \centering
    \begin{subfigure}[b]{0.50\textwidth}
        \includegraphics[width=\linewidth]{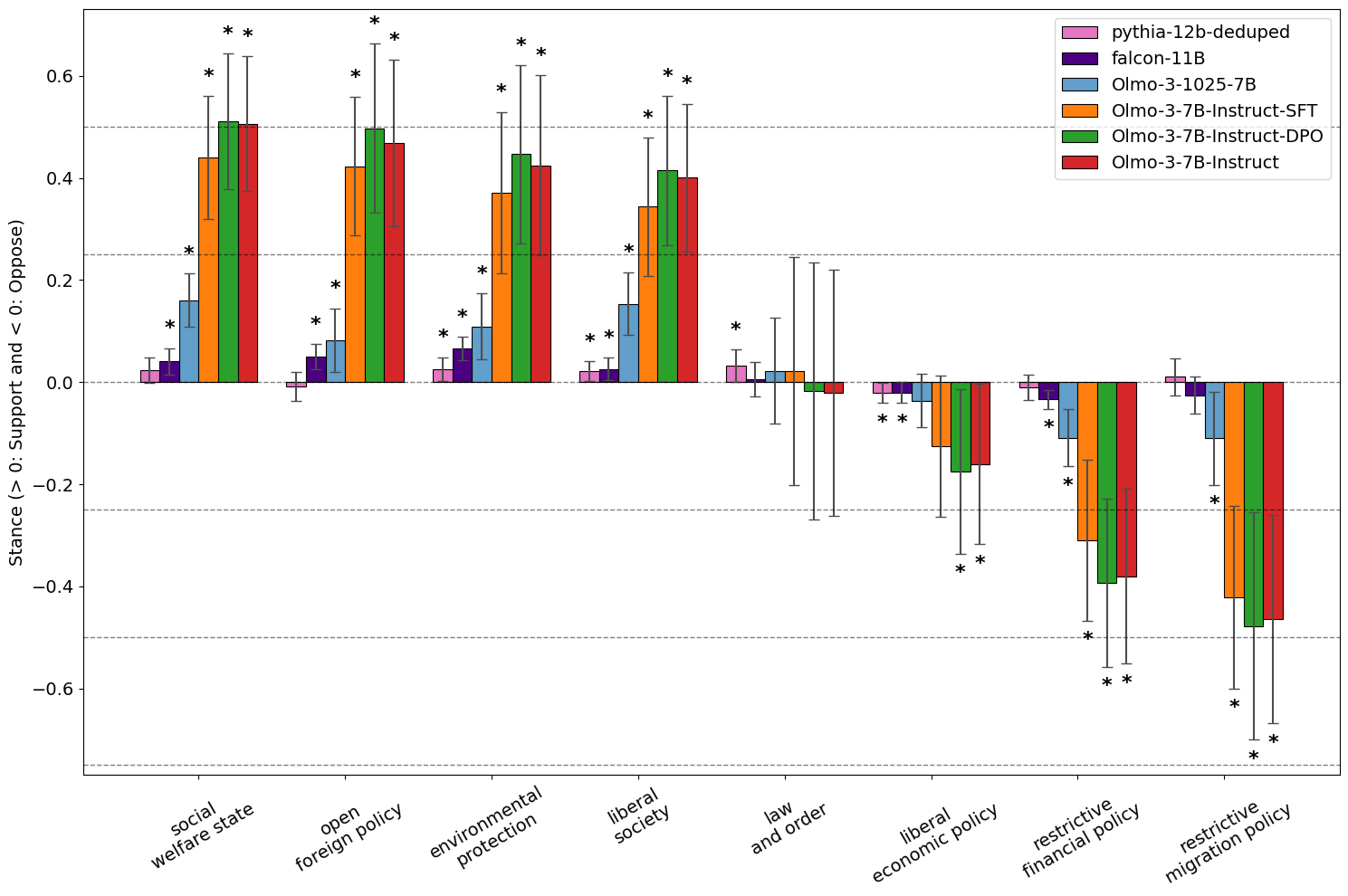}
        \caption{Stance in models}
        \label{fig:olmo3-stances}
    \end{subfigure}%
    \hfill
    \begin{subfigure}[b]{0.50\textwidth}
        \includegraphics[width=\linewidth]{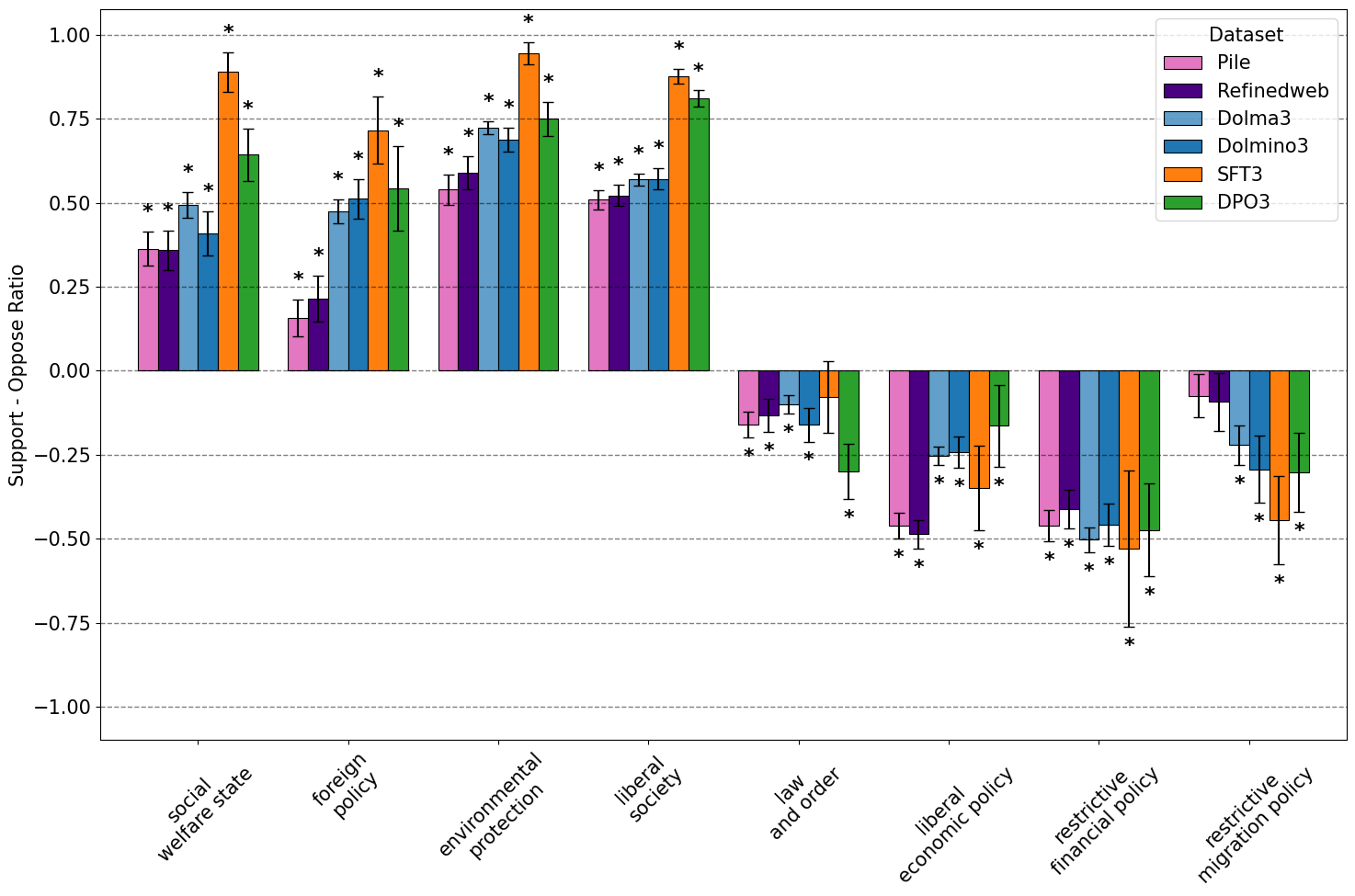}
        \caption{Stances in documents from training data}
        \label{fig:stance3-documents}
    \end{subfigure}
    \caption{Stances of the models at different training stages and in the documents from the training data across policy issues. Colors of the bars match models and datasets used to train them at that stage. \textsc{Dolma} and \textsc{Dolmino} are in blue because they are both used in the base model. Error bars show the 95\% confidence intervals for the stance scores. Statistical significance was assessed with a one-sample two-proportion z-test (testing whether the support proportion differs from 0), Holm-corrected across all datasets/models$\times$policy bars.}
    \label{fig:stance-documents-model}
\end{figure*} 

\section{Results}
In this section, we present the results of the classifiers and the correlational analysis between training data and model behavior. 

\subsection{Politically-engaged documents}
Overall, the results show that the pre-training datasets contain a higher proportion of politically-engaged documents, with an average of 5.4\% ($\pm$0.71), in comparison with the post-training datasets with an average of 2.46\% ($\pm$0.16). Table \ref{tab:olmo-datasets} shows that \textsc{Pile} has the highest share of politically engaged documents, with 1.75\% right-leaning and 4.46\% left-leaning ones, while \textsc{RefinedWeb} has the lowest, with 3.13\% left- and 1.38\% right-leaning documents. Second and third in ranking are \textsc{Dolma3} and \textsc{Dolmino3} with similar percentages of 3.84\% left- vs.\ 2.17\% right- and 3.06\% left- vs.\ 1.85\% right-leaning documents, respectively. Among the post-training datasets, \textsc{SFT3} has the highest proportion, with 2.27\% left- and 0.36\% right-leaning and documents, while \textsc{DPO3} has 1.88\% left- and 0.42\% right-leaning documents. 

The results thus show a consistently higher proportion of left-leaning documents across datasets with an average ratio of left- to right-leaning documents of 3.16 ($\pm$1.67). This finding aligns with the results of previous studies related to political opinions showing that the content generated by models reflects a generally more left- and progressive leaning orientation \citep{ceron-etal-2024-beyond,motoki2023more,nikolaev2026llmsreadyhardchoices}. 

\subsection{Topics}
Topics prevalent in left- and right-leaning documents in the surveyed datasets and keywords for the topics are reported in Tables \ref{tab:top-topics} and \ref{tab:common-topics-keywords}, respectively, in Appendix \ref{appendix:topic-analysis}. The results show that there is a large degree of overlap, with many documents focusing on \enquote{across-the-aisle} issues such as the workings of the federal government, climate protection, and the relationship with China, there are obvious topical differences, with left-leaning documents focusing on the rights of the racial or LGBTQ minorities, renewable energy and affordable housing, gender equality, representation of women and people of color in the media, and agriculture in California. Right-leaning texts, by contrast, pay a lot of attention to the problems of net neutrality, markets, taxation, Christianity, gun laws and law enforcement, and legal issues around abortion.

These tendencies are not reflected uniformly in the datasets, with Dolma3 and Dolmino3 having very similar topic distributions, SFT3 and Pile demonstrating a slightly divergent set of topics, and DPO3 and RefinedWeb only having documents that belong to the common left-right pool under this analysis, with left- and right-specific topics not making it to the top 20.

\subsection{Political stances reflected in LLMs}
\label{subsec:models-stance}

Figure \ref{fig:olmo3-stances} illustrates the results of the stances reflected in models. Overall, responses show a favorable attitude towards
left-leaning policies (\textit{Expanded Social Welfare State}, \textit{Open Foreign Policy}, \textit{Expanded Environmental Protection}, and \textit{Liberal Society}), 
and opposition towards \textit{Liberal Economic Policy}, \textit{Restrictive Migration Policy} and \textit{Restrictive Financial Policy}.

The stances across all \textsc{Olmo3} checkpoints follow the same direction differing only in the magnitude. As observed in previous studies, base models demonstrate a rather weak directional signal given their inability to follow prompt instructions even if the prompt is formatted to be continued \citep{ceron-etal-2024-beyond}. \textsc{Pile} does not have a statistically significant stance in 4 and \textsc{RefinedWeb} and \textsc{Olmo3} base in 2 out of 8 policy issues. The direction of the stance, though, is very similar across 
models except for \textsc{Pythia-12b}, which alone has a slight tendency to support \textit{Restrictive Migration Policy} and is rather in between support and oppose for \textit{Open Foreign Policy} and \textit{Restrictive Financial Policy}. Among policy issues, \textit{Law and Order} stands out because of the very weak stance signal across models except for \textsc{Pile}, suggesting that models do not have a strong preference in this domain.

\subsection{Stances in the training data}
\label{subsec:stance-training-data}

Figure \ref{fig:stance3-documents} summarizes the aggregated stances of documents in the training data (see Table \ref{tab:results-stance-classification-docs} for the proportion of support and oppose documents). The direction of the stance is the same across all pre- and post-training datasets, changing only in the magnitude of some policy issues. To test whether these magnitude differences are statistically robust, we compare each post-training dataset against the pooled pre-training baseline (\textsc{Pile}+\textsc{RefinedWeb}+\textsc{Dolma3}+\textsc{Dolmino3}) using a two-proportion z-test with Holm correction ($\alpha$=0.05, Table \ref{tab:sft-dpo-vs-baseline}). \textsc{SFT3} shows a significantly stronger stance for five of eight policy issues (\textit{Expanded Social Welfare State}, \textit{Open Foreign Policy}, \textit{Expanded Environmental Protection}, \textit{Liberal Society}, \textit{Restrictive Migration Policy}), 
confirming its larger difference in magnitude is not sampling noise. \textsc{DPO3} reaches significance for only three of them instead: \textit{Expanded Social Welfare State}, \textit{Liberal Society}, and \textit{Law and Order}. 

\textit{Law and Order} is the policy issue showing the weakest stance signal across datasets except for \textsc{DPO3}. A full pairwise comparison across all six datasets confirms this (Table \ref{tab:law-and-order-all-pairs}): \textsc{DPO3} differs significantly from \textsc{Dolma3}, \textsc{RefinedWeb}, \textsc{SFT3}, and \textsc{Pile} (p-Holm$<$0.05), with only \textsc{Dolmino3} falling short of significance, while none of the remaining ten pairwise comparisons reach significance. This indicates that \textsc{Dolma3}, \textsc{Dolmino3}, \textsc{RefinedWeb}, \textsc{Pile}, and \textsc{SFT3} have a statistically indistinguishable stance on \textit{Law and Order}, with \textsc{DPO3} alone driving the difference from this otherwise uniformly weak stance signal.

Beyond the pre- vs. post-training comparison, we test whether pre-training datasets differ from one another, since Figure \ref{fig:stance3-documents} suggests that \textsc{Pile} and \textsc{RefinedWeb} diverge from \textsc{Dolma3} and \textsc{Dolmino3} on some issues. We compare the proportion of \textit{support} documents (out of \textit{support}+\textit{oppose}) between \textsc{Pile}/\textsc{RefinedWeb} and \textsc{Dolma3}/\textsc{Dolmino3} pairwise, using a two-proportion z-test with Holm correction ($\alpha$=0.05; Table \ref{tab:pile-refinedweb-vs-dolma-dolmino}). \textsc{Pile} and \textsc{RefinedWeb} differ significantly from both \textsc{Dolma3} and \textsc{Dolmino3} on \textit{Open Foreign Policy} and \textit{Liberal Economic Policy}, consistent with the roughly double stance magnitude seen in Figure \ref{fig:stance3-documents} for the latter. \textsc{Pile} additionally differs from \textsc{Dolma3} on \textit{Expanded Social Welfare State} and \textit{Liberal Society}, and from \textsc{Dolmino3} on \textit{Restrictive Migration Policy}, suggesting \textsc{Pile} is somewhat more diverging than \textsc{RefinedWeb}. \textit{Law and Order} and \textit{Restrictive Financial Policy} show no significant differences across any pair, consistent with their weak, uniform stance signal. Overall, while the four pre-training datasets broadly agree in stance direction (Table \ref{tab:pearson-correlation-model-dataset}, average 
Pearson $r$=0.96), \textit{Open Foreign Policy} and \textit{Liberal Economic Policy} show robust, dataset-specific differences in stance magnitude.

\begin{table}[]
    \centering
    \footnotesize
    \begin{tabular}{llrrrr}
    \toprule
    Model & Dataset & $r$  & CI low & CI high & SE\\
    \midrule
    O-base & Dolma & 0.92* & 0.62 & 0.99 & 0.04 \\
    O-base & Dolmino & 0.91* & 0.59 & 0.98 & 0.05 \\
    O-SFT & SFT3 & 0.96* & 0.80 & 0.99 & 0.02 \\
    O-DPO & DPO3 & 0.93* & 0.67 & 0.99 & 0.04 \\
    Instruct & DPO3 & 0.93* & 0.67 & 0.99 & 0.04 \\
    Falcon & Refined & 0.88* & 0.46 & 0.98 & 0.08 \\
    Pythia & Pile & 0.66 & -0.09 & 0.93 & 0.28 \\
    \bottomrule
    \end{tabular}
    \caption{Pearson correlation between the models' stances at different training stages and the stances of the training documents. *$=p$-value $<0.05$. SE indicates Jackknife standard errors with leave-one-policy-out.}
    \label{tab:pearson-correlation-model-dataset}
\end{table}

\subsection{Correlation between data and LLMs}
\label{subsec:correlation}

Now that we have quantified the political stances in model outputs and in training data separately, we compute Pearson correlations to investigate the association between them. Overall, we find that the stances of the training documents are strongly correlated with model stances, with an average Pearson $r$ of 0.88 across model–dataset pairs (see Table~\ref{tab:pearson-correlation-model-dataset}), all with tight bootstrap confidence intervals and small Jackknife standard errors (leave-one-policy-out), except for \textsc{Pile}, which is not significant because of the number of invalid answers (Cf. Figure \ref{fig:num-nas}) and correspondingly has the widest confidence interval ($r$=0.66, CI [-0.09, 0.93], SE=0.28). Pearson correlations do not increase considerably with subsequent training stages compared to the base model. In \textsc{OLMO3} models, $r$ varies from 0.91 (CI [0.59, 0.98], SE=0.05) between \textsc{Dolmino3} and \textsc{OLMO3-7B} base to 0.96 (CI [0.80, 0.99], SE=0.02) between \textsc{SFT3} and \textsc{OLMO3-7B-SFT}, with \textsc{DPO3} correlating at $r$=0.93 (CI [0.67, 0.99], SE=0.04) with both \textsc{OLMO3-7B-DPO} and \textsc{OLMO3-7B-Instruct}.

\section{Discussion}
\label{sec:discussion}

We now discuss how our findings shed light on the four research questions (R1–R4) introduced in \S\ref{sec:intro}. 

\noindent\textbf{R1: Imbalance in training data.} Across all analyzed datasets, left-leaning documents consistently outnumber right-leaning ones, with ratios ranging from 1.6 to 6.2 depending on the corpus. This skew is corroborated by an independent and classifier-free estimate in the pre-training datasets. Given the web domain information is present in the metadata, we compute the proportion of left-leaning and right-leaning news articles based on the political leaning of news outlets (Cf. Appendix \ref{sec:classifier-free}). This shows the same directional imbalance across \textsc{Dolma3}, \textsc{Dolmino3}, and \textsc{RefinedWeb}. 

\noindent\textbf{R2: Variance across pre-training datasets.} As Table \ref{tab:final-dataset-comparison} in Appendix shows, even though the largest amount of the pretraining data comes from web-scraped documents from datasets such as C4 \citep{raffel2020exploring}, DCLM \citep{li2024datacomp}, and FineWeb \citep{penedo2024fineweb}, filtering and curation strategies appear to have only a limited effect on the overall political content present in training data. Our analysis across models with non-overlapping datasets shows strong aggregate correlation across corpora (average $r$=0.96; Table \ref{tab:pearson-corr-datasets}), suggesting that curation largely preserves the broad political orientation of a corpus. In terms of magnitude of stances, however, pairwise testing (Table \ref{tab:pile-refinedweb-vs-dolma-dolmino}) shows that \textsc{Pile} and \textsc{RefinedWeb}, the least curated datasets, differ significantly from \textsc{Dolma3}/\textsc{Dolmino3} on \textit{Open Foreign Policy} and \textit{Liberal Economic Policy}. Curation therefore appears to keep direction the same while still shifting stance magnitude on some issues, which better explains the high similarity in political stances across model families observed here and in prior work \citep{ceron-etal-2024-beyond,weeber-etal-2026-political}.

\noindent\textbf{R3: Correlation between training data and model behavior.} The political stances present in training data are highly correlated with those expressed by models across policy issues, with an average Pearson correlation of $r$=0.88. This result indicates that model behavior closely reflects the distribution of political content in the underlying data. This relationship appears to scale with signal strength: \textsc{SFT3}, the dataset with the most policy issues showing a significantly amplified stance relative to the pre-training baseline (Table \ref{tab:sft-dpo-vs-baseline}), also yields the highest data–model correlation of any dataset ($r$=0.963), suggesting that a clearer stance signal in the data translates into tighter alignment with model outputs rather than the aggregate correlation being driven by a few outlier issues. However, this is to be taken with a grain of salt because base models yield more inconsistent answers than post-trained models as discussed in \S~\ref{subsec:correlation}. Another evidence to this hypothesis is the fact that the policy issue (\textit{Law and Order}) with the weakest stance in the training data is also the issue with the weakest stance reflected in the responses of the models as shown in Figure \ref{fig:olmo3-stances}. While this correlation does not establish causality, it provides evidence that training data composition is an important factor correlating with model outputs. This is consistent with the findings by \citet{xu-etal-2025-better}, who show a strong correlation between political stances related to the US court issues and the training data. 

\noindent\textbf{R4: (Dis)similarities across training stages.} The direction of political stances is the same both across model checkpoints and across training datasets, with \textsc{OLMO3-base} already exhibiting the same directional pattern found in the post-trained checkpoints (Figure \ref{fig:olmo3-stances}). While \textsc{SFT3} and \textsc{DPO3} contain substantially less politically-engaged content overall than pre-training corpora (2.46\% vs. 5.4\% on average), the political content they do contain is proportionally more skewed than any pre-training dataset, with L/R ratios of 6.29 and 4.47 respectively compared to 1.66–2.55 for pre-training corpora (Table \ref{tab:olmo2-datasets}). Second, post-training has a higher magnitude in most policy issues: \textsc{SFT3} significantly amplifies the stance relative to the pooled pre-training baseline on five of eight policy issues, and \textsc{DPO3} on three of eight, with both stages concentrating this amplification on the same subset of issues (Expanded Social Welfare State, Liberal Society) while stance magnitude is higher in pre-training only in Liberal Economic Policy. Third, despite this added training stage, the correlation between training-data stance and model behavior increases only modestly, from r=0.91 between \textsc{dolmino3} and \textsc{OLMO3-base} to r=0.96 between \textsc{SFT3} and the \textsc{OLMO3-SFT} (\S~\ref{subsec:correlation}), suggesting that the alignment between data composition and model output is already established during pre-training. We note that this analysis of training-stage trajectories relies on the single model family \textsc{OLMO3} for which complete pre- and post-training data are available.


\section{Related work}


\citet{elazar2024whats} propose a search and count method to process and analyze the bulk of the pre-training data in an accessible way. \citet{piktus-etal-2023-roots} make available a search tool to explore ROOTS, the dataset used to train the open source model BLOOM, with fuzzy or exact matches, while \citet{piktus-etal-2023-gaia} propose a search engine 
to understand datasets before even using them for training. Other works have instead 
analyzed the effect of data on model behavior, for example in the context of fact checking \citep{akyurek-etal-2022-towards}, looked at the correlation between model performance and data quality \citep{longpre-etal-2024-pretrainers}, and estimated the similarity of generated text to the training data \citep{mccoy-etal-2023-much}. 

The paper by \citet{xu-etal-2025-better} use \citeauthor{elazar2024whats}'s (\citeyear{elazar2024whats}) search tool to extract documents that contain terms related to 32 U.S.\ Supreme Court cases on topics including abortion and voting rights. \citet{waight2026state} subsequently investigates how the contamination of Chinese state-media influence in an open dataset. \citet{10.1145/3773966.3777990} have investigated political content in one training dataset (C4) through keywords on very specific topics. Our work differs from their approach 
in that we (i)~evaluate political content in broader policy issue categories in a way more applicable for cross-country comparisons, 
(ii)~provide a wider coverage of pre-training corpora aggregated in 
\textsc{Dolma3} and \textsc{Dolmino3} datasets, and
(iii)~shed light on post-training, which is highly relevant to how models eventually behave in user interactions given their steerability in that phase \citep{stammbach-etal-2024-aligning,weeber-etal-2026-political}. 

The work investigating political opinions in LLMs is shared between the computational social science and computer science literature. Many studies have conducted a descriptive analysis of the political leaning present in LLMs \citep[e.g.,][]{motoki2023more, hartmann2023political, Rutinowski2024, rottger-etal-2024-political, ceron-etal-2024-beyond,rottger2025issuebench}. Overall, recent results suggest that large commercial LLMs reflect a consistent left-leaning opinions, whereas political leanings for smaller open-source models seem to be more mixed and less consistent in their stance.

Another line of research has looked into the steerability of aligning large language models with specific political opinions \citep{jiang-etal-2022-communitylm, feng2023pretraining, stammbach-etal-2024-aligning} and has found that it is possible to slightly change the political leaning, steering it to be less left-leaning. While the bulk of work on LLM stances focuses on the US, similar results have been observed from the point of view of Swiss politics
\citep{stammbach-etal-2024-aligning}, and it has been observed that steering the leaning in one language also impacts political beliefs in other languages \citep{weeber-etal-2026-political}.

Another set of studies has focused on the impact of political leanings in LLMs on human political beliefs \citep{potter-etal-2024-hidden, doi:10.1073/pnas.2403116121, Bai2025}. They show that LLMs are (hidden) persuaders, as they can influence political beliefs and potentially impact voting behavior.
More broadly, these studies contribute to the literature on how LLMs can influence human decision-making \citep{doi:10.1073/pnas.2017548118, sharma2024generativeechochambereffects}. 

\section{Conclusion} 

This paper presents the first large-scale, data-centric analysis that quantifies political content in training data and links the political composition of pre- and post-training corpora to the political behavior of the resulting LLMs. 

While our paper quantifies the extent of the skew of political leaning and political stances in the training data, it does not argue for equal balance across the spectrum for mitigation in the sense of correction. Reaching real political neutrality in LLMs is impossible given that giving equal opportunities for all different opinions to flourish might go as far as violating constitutions or human rights (e.g. apology for Nazism). Our aim instead is to understand the relation between training data and model behavior as a way of providing transparency to the data that is being fed into LLMs and its correlation with model behavior. 

Future work should explore to what extent modifying the training data might have an effect on the different political stances of models, placing particular attention in the relation between political stances and factuality and objectivity. In this paper, we show that model behavior highly correlates with its training data. We argue therefore for transparency about the composition of training data, so that stakeholders can on the one hand make informed decisions about their use, research purpose, policy-making decisions, and on the other hand, monitor model behavior. What matters is not that the data mirror some notion of ideological neutrality given that it might be an unreachable goal, but that its composition is documented and legible, so that any resulting model behavior can be correctly attributed and audited. 

\section*{Limitations}

These findings should be read alongside some limitations. Our left–right framing, while methodologically standard, necessarily simplifies a richer ideological space and discards centrist or ambiguously political content. Moreover, our classifiers, though validated, show asymmetric performance across left- and right-leaning categories. As shown in \S\ref{subsec:document-classification}, models struggle the most with classifying left-leaning documents, often mislabeling them as "neutral" -- potentially underestimating the real proportion of left-leaning documents in the training dataset. Since we deliberately selected model families with non-overlapping corpora to rule out shared-data artifacts, only \textsc{OLMO3} has all training stages, limiting our ability to generalize the base-to-post-training trajectory across architectures. Moreover, the strong correlations we report are associative rather than causal: they establish that data and model stance move together, not that altering the data would proportionally shift model behavior. Finally, we evaluate models and training data documents in English. Extending this analysis beyond English to multilingual corpora and non-Western political spectra would also test the generality of our findings.

\subsection*{Ethical Considerations}

With this study, we quantify the correlation of training data and model behavior, raising both ethical concerns and societal implications. We are doing it with the best intention of understanding models. However, the ability to manipulate models' political leanings can, in the hands of ill-intentioned actors, serve malicious purposes to deliberately change the behavior of LLMs, in order to automate the generation of misinformation or propaganda at scale. 

Our intentions are that advancing research in this area can provide valuable insights for policymakers in the regulation of AI systems, as a deeper understanding of their mechanisms facilitates the development of more effective regulatory frameworks. For example, given the importance of pre-training data in models' knowledge, policymakers may demand a greater level of detail in the documentation of the data used for pre-training LLMs, and in particular, explanation of specific choices taken when curating training data. This is strongly linked to accountability. Making design choices more explicit holds AI providers accountable for what they do. All in all, we aim to contribute to the responsible research trajectory by empirically correlating political stances with pre-training data, thereby informing future efforts toward more pluralistic, transparent, and trustworthy AI systems.


\bibliography{custom}

\appendix

\section{Dataset Comparison}
\label{appendix:dataset-comparison}

Table \ref{tab:final-dataset-comparison} shows the datasets present across pre-training corpora. 

All the data used from OLMO2 and OLMO3 is under Open Data Commons Attribution License v1.0 (ODC-By). It is intended for research and educational use.

\begin{table*}[t]
\centering
\tiny
\resizebox{\linewidth}{!}{%
\begin{tabular}{lccccccc}
\toprule
Dataset \textbackslash~Model & OLMO2 & Marin-8B & SmolLM3 & Apertus & Falcon2
& Pythia & EuroLLM \\
\midrule
Dolma & \cmark & \cmark & \xmark & \xmark & \xmark & \xmark & \xmark \\
Dolmino & \cmark & \cmark & \cmark & \xmark & \xmark & \xmark & \xmark \\
DCLM-Baseline & \cmark & \cmark & \cmark & \xmark & \xmark & \xmark & \xmark \\
DCLM-Edu & \xmark & \xmark & \xmark & \cmark & \xmark & \xmark & \xmark \\
Overlap DCLM & \cmark & \cmark & \cmark & \cmark & \xmark & \xmark & \xmark \\
FineWeb & \cmark & \xmark & \xmark & \xmark & \xmark & \xmark & \xmark \\
FineWeb-2 (Mlingual) & \xmark & \xmark & \xmark & \cmark & \xmark & \xmark & \xmark \\
FineWeb-2-HQ & \xmark & \xmark & \xmark & \cmark & \xmark & \xmark & \xmark \\
FineWeb-Edu & \xmark & \xmark & \cmark & \cmark & \xmark & \xmark & \cmark \\
FineWeb-HQ & \xmark & \xmark & \cmark & \cmark & \xmark & \xmark & \xmark \\
Overlab FINEWEB & \cmark & \xmark & \cmark & \cmark & \xmark & \xmark & \cmark \\
Pile & \xmark & \xmark & \xmark & \xmark & \xmark & \cmark & \xmark \\
Wikipedia & \cmark & \cmark & \cmark & \xmark & \xmark & \xmark & \cmark \\
ArXiv & \cmark & \cmark & \cmark & \xmark & \cmark & \xmark & \xmark \\
Books & \cmark & \xmark & \xmark & \xmark & \cmark & \xmark & \cmark \\
peS2o & \cmark & \cmark & \cmark & \xmark & \xmark & \xmark & \xmark \\
FLAN & \cmark & \cmark & \cmark & \xmark & \xmark & \xmark & \xmark \\
Nemotron-CC & \xmark & \cmark & \xmark & \xmark & \xmark & \xmark & \xmark \\
Proof-Pile-2 & \xmark & \cmark & \xmark & \xmark & \xmark & \xmark & \xmark \\
SmolLM corpus & \xmark & \xmark & \cmark & \xmark & \xmark & \xmark & \xmark \\
RefinedWeb & \xmark & \xmark & \xmark & \xmark & \cmark & \xmark & \xmark \\
Reddit & \xmark & \xmark & \xmark & \xmark & \cmark & \xmark & \xmark \\
HackerNews & \xmark & \xmark & \xmark & \xmark & \cmark & \xmark & \xmark \\
CC-NEWS & \xmark & \xmark & \xmark & \xmark & \xmark & \xmark & \xmark \\
CC-MAIN-2024-33 through CC-MAIN-2025-13 & \xmark & \xmark & \xmark & \xmark & \xmark & \xmark & \xmark \\
Multilingual data & \xmark & \xmark & \xmark & \xmark & \cmark & \xmark & \xmark \\
Scientific PDFs & \xmark & \xmark & \xmark & \xmark & \xmark & \xmark & \xmark \\
Apollo & \xmark & \xmark & \xmark & \xmark & \xmark & \xmark & \cmark \\
Cosmopedia & \xmark & \xmark & \cmark & \xmark & \xmark & \xmark & \cmark \\
RedPajama-Data-v2 & \xmark & \xmark & \xmark & \xmark & \xmark & \xmark & \cmark \\
HPLT (MLingual) & \xmark & \xmark & \xmark & \xmark & \xmark & \xmark & \cmark \\
MADLAD-400 (MLingual) & \xmark & \xmark & \xmark & \xmark & \xmark & \xmark & \cmark \\
CulturaX (MLingual) & \xmark & \xmark & \xmark & \xmark & \xmark & \xmark & \cmark \\
 mC (MLingual) & \xmark & \xmark & \xmark & \xmark & \xmark & \xmark & \cmark \\
\bottomrule
\end{tabular}
}
\caption{Comparison of pretraining datasets used across models. Mlingual means that the dataset has been built to be multilingual. \textit{Overlap FineWeb} means that the there is some overlap with any of the versions of FineWeb. Note that Dolma and Dolmino encompass portions of other datasets such as DCLM and FineWeb. The Math, code, and reasoning datasets are not included in the list.}
\label{tab:final-dataset-comparison}
\end{table*}

\section{Left-right-neutral classification}

\subsection{News Article Dataset}
\label{appendix:news-article-dataset}

Each news article was first manually annotated by two expert annotators recruited on Upwork. The requirement was to hold a postgraduate degree in relevant subjects (e.g., political science). After that, a third annotator, who is an undergraduate student, adjudicated the disagreements between the two expert annotators. The annotators labeled the articles with labels ranging from 1 to 7 with 1 being very left and 7 being very right. We mapped the labels between 3.5 and 4.5 as neutral, between 1 and 3 as left and between 5 and 7 as right. The annotators were given the instruction to consider partisan use of language, political alignment, and the issues covered in the articles, given that some issues are known to attract more attention by left- resp.\ right-leaning politicians and activists. The mapped dataset contains 2659 neutral articles, 1148 left-leaning and 627 right-leaning articles. We undersample the neutral documents to be the same size as the left-leaning documents for our validation set. 

\subsection{Political News Classification}
\label{appendix:left-right-classification}

Table \ref{tab:prompt-templates-results} shows the results of the evaluated classifiers and prompts for the political leaning task. 
Table \ref{tab:classification-results-ensemble} shows the results of the political leaning classification with the final chosen best prompt and model on the entire dataset. Tables \ref{tab:classification-results-qwen} and  \ref{tab:classification-results-gemma} show the results of the classifiers alone. 

We ran the prompts on a single 80GB GPU. On average, it took 24 hours to process every 100k documents, as a rough estimate, given that the time varies according to the document's size.

\begin{table*}[]
    \centering
    \small
    \begin{tabular}{lrrrrrr}
    \toprule
    model & Qwen3.5-27B & Qwen3.6-27B & gemma-4-31B-it & gpt-5.4-mini & Llama-3.3-70B & f1-models \\
    \midrule
    Prompt 2 & 0.711 & 0.730 & 0.706 & 0.742 & 0.708 & 0.714 \\
    Prompt 13 & 0.708 & 0.721 & 0.701 & 0.734 & 0.713 & 0.711 \\
    Prompt 1 & 0.703 & 0.721 & 0.706 & 0.729 & 0.713 & 0.711 \\
    Prompt 6 & 0.678 & 0.662 & 0.629 & 0.491 & 0.720 & 0.672 \\
    Prompt 5 & 0.730 & 0.720 & 0.697 & 0.716 & 0.514 & 0.665 \\
    Prompt 8 & 0.732 & 0.714 & 0.672 & 0.722 & 0.533 & 0.663 \\
    Prompt 11 & 0.717 & 0.709 & 0.679 & 0.728 & 0.536 & 0.660 \\
    Prompt 9 & 0.719 & 0.701 & 0.684 & 0.728 & 0.528 & 0.658 \\
    Prompt 12 & 0.719 & 0.706 & 0.670 & 0.732 & 0.537 & 0.658 \\
    Prompt 4 & 0.717 & 0.697 & 0.681 & 0.702 & 0.536 & 0.658 \\
    Prompt 7 & 0.713 & 0.701 & 0.674 & 0.727 & 0.542 & 0.658 \\
    Prompt 10 & 0.709 & 0.706 & 0.674 & 0.727 & 0.538 & 0.657 \\
    Prompt 14 & 0.716 & 0.699 & 0.681 & 0.725 & 0.532 & 0.657 \\
    Prompt 3 & 0.716 & 0.701 & 0.661 & 0.728 & 0.539 & 0.654 \\
    \hline
    ModernBERT-base & Fine-tuned & 0.50 &  &  &  & \\
    ModernBERT-large & Fine-tuned & 0.59 &  &  &  & \\
    \bottomrule
    \end{tabular}
    \caption{Results of the prompt with open and closed models. Avg. f1-macro is the average f1-macro scores across open models only.}
    \label{tab:prompt-templates-results}
\end{table*}

\begin{tcolorbox}[
  title=Best Prompt with Qwen3.5-27B and Gemma4-31B (quantized) for political leaning classification,
  colback=gray!10,
  breakable
]
\begin{verbatim}
You are an expert political analyst 
tasked with classifying documents by 
their political slant based on three 
key factors: language choices, 
political positions, and issue 
coverage. Please analyze the 
following document and 
classify its political slant 
as LEFT, NEUTRAL, or RIGHT.

Consider these three factors in 
your analysis:

Language Analysis: Look for 
partisan terminology
- Right-leaning terms: "death 
tax," "job creators," "government 
overreach," "pro-life"
- Left-leaning terms: "estate 
tax," "wealthy elite," "corporate 
welfare," "pro-choice"

Political Position: Assess 
alignment with party positions
- Right-leaning: supports 
conservative policies (lower 
taxes, deregulation, traditional 
values)
- Left-leaning: supports progressive 
policies (social programs, 
regulation, social justice)

Coverage Selection: Evaluate what 
aspects are emphasized or omitted
- Does the document present only 
one side of a multi-faceted issue?
- Are counterarguments or alternative 
perspectives acknowledged?

Document to analyze:
{document}

Classification: Based on your 
analysis, classify this document as:
1. LEFT: Shows clear bias toward 
Democratic/progressive viewpoints
2. NEUTRAL: Presents balanced 
coverage without clear partisan lean
3. RIGHT: Shows clear bias toward 
Republican/conservative viewpoints

Respond with only the digit (1, 2, or 
3) and no explanation.
ANSWER:
"""
\end{verbatim}
\end{tcolorbox}

\begin{table}[t]
\centering
\small
\begin{tabular}{lrrrr}
\toprule
 & precision & recall & f1-score & support \\
\midrule
left & 0.862 & 0.506 & 0.638 & 1148 \\
neutral & 0.610 & 0.897 & 0.726 & 1148 \\
right & 0.813 & 0.727 & 0.768 & 627 \\
accuracy & 0.707 & 0.707 & 0.707 & 1 \\
macro avg & 0.762 & 0.710 & 0.711 & 2923 \\
weighted avg & 0.753 & 0.707 & 0.700 & 2923 \\
\bottomrule
\end{tabular}   
\caption{Results of the left-neutral-right classification with the best prompt and Qwen3.5-27B on the \texttt{NEWSLEAN} dataset.}
\label{tab:classification-results-qwen}
\end{table}

\begin{table}[t]
\centering
\small
\begin{tabular}{lrrrr}
\toprule
 & precision & recall & f1-score & support \\
\midrule
left & 0.865 & 0.502 & 0.635 & 1148 \\
neutral & 0.591 & 0.910 & 0.717 & 1148 \\
right & 0.871 & 0.681 & 0.765 & 627 \\
accuracy & 0.701 & 0.701 & 0.701 & 1 \\
macro avg & 0.776 & 0.698 & 0.706 & 2923 \\
weighted avg & 0.759 & 0.701 & 0.695 & 2923 \\
\bottomrule
\end{tabular}
\caption{Results of the left-neutral-right classification with the best prompt and Gemma4-31B on the \texttt{NEWSLEAN} dataset.}
\label{tab:classification-results-gemma}
\end{table}


\begin{figure}[t]
    \centering
    \includegraphics[width=0.8\linewidth]{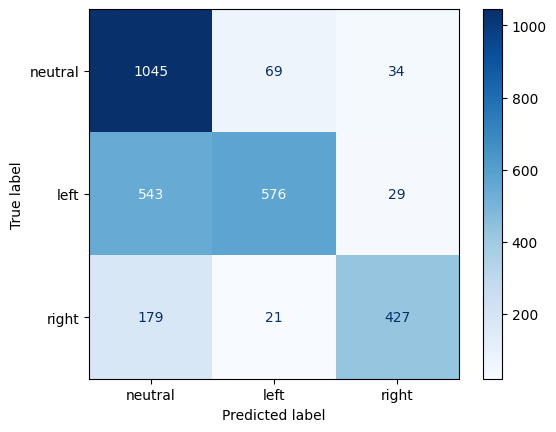}
    \caption{Confusion matrix of the left-neutral-right classification with the ensemble setup with Qwen3.5 and Gemma4. The ensemble classification considers documents as left or right whenever one of the two models labeled it as left or right and the other neutral. Whenever one model classifies as left and the other right, the tie breaker is the Gemma4 model as it yielded better results in the \textsc{Newslean} dataset.}
    \label{fig:cm-classification}
\end{figure}

\subsection{Validation of the leaning classifier}
\label{sec:validation}

Table \ref{tab:directions} shows the conversion of stance labels to political leaning. Given that each document has 8 labels for each of the policy issue, we make sure to include documents that are only labeled "right" or "left" across all policy issues.  We do not include “neutral” labels in this dataset because we cannot guarantee that what is neither support or oppose of the eight policies does not contain a political ideology. 

Table \ref{tab:classification-results-ensemble} shows the results of the classifier on \textsc{stancepol}. 

\begin{table}[ht]
\centering
\small
\begin{tabular}{lcc}
\toprule
Direction & Support & Oppose \\
\midrule
open\_foreign\_policy & left & right \\
liberal\_economic\_policy & right & left \\
law\_and\_order & right & left \\
restrictive\_migration\_policy & right & left \\
expanded\_environ\_protection & left & right \\
expanded\_social\_welfare\_state & left & right \\
liberal\_society & left & right \\
restrictive\_financial\_policy & right & left \\
\bottomrule
\end{tabular}
\caption{Directions and their corresponding leanings when converting \textsc{stancepol} into \textsc{stancepol-lean}.}
\label{tab:directions}
\end{table}


\subsection{Classifier-free estimation}
\label{sec:classifier-free}

We also estimate the proportion of news articles from the left- and right-leaning news outlets in pre-training data as another way of verifying the number imbalance in the data. 

We compute the proportions of left- and right-leaning news outlets found in the origin domains of the pre-training datasets. This computation is based on the domain-level annotations in \textsc{newslean}. An outlet is labelled left-leaning if more of its articles in \textsc{newslean} are annotated as left than as right, and right-leaning otherwise; articles with a neutral slant (4.0) are ignored. This yields 94 outlets, 74 left- and 20 right-leaning. We then extract the root domain of each document URL in the pre-training data (e.g., \texttt{cnn.com}) and match it to these outlets. To focus on politically slanted content, we restrict the analysis to the documents our classifier assigns to the left or right class. SFT3, DPO3, and Pile are excluded because they do not include source URLs. 

We estimate the shares over all documents in each dataset (left, right, and neutral). To account for the fact that documents from the same outlet are not independent, we compute 95\% confidence intervals with a cluster bootstrap that resamples root domains with replacement (1{,}000 resamples), and test the one-sided hypothesis that left-leaning outlets are more prevalent than right-leaning ones. Documents from left-leaning outlets account for 1.34\% of RefinedWeb, 1.09\% of Dolma3, and 0.10\% of Dolmino3, compared to 0.22\%, 0.20\%, and 0.02\% for right-leaning outlets. Among documents matched to a left- or right-leaning outlet, the left-leaning share is 85.9\%, 84.6\%, and 79.9\%, respectively. The excess of left- over right-leaning outlet documents is statistically significant in all three datasets ($p < 0.001$; the one-sided 95\% lower bound on the difference is 0.74, 0.52, and 0.03 percentage points). These intervals reflect document sampling variability only and take the outlet-level labels from \textsc{newslean} as given.

\section{Topic Modeling Analysis}
\label{appendix:topic-analysis}

Tables \ref{tab:top-topics} and \ref{tab:common-topics-keywords} show the results of the topic modeling analysis. 

\begin{table*}[t]
\footnotesize
\centering
\begin{tabular}{@{}llll@{}}
\toprule
\textbf{Dataset} & \textbf{Common topics} & \textbf{Left only} & \textbf{Right only} \\ \midrule
Dolma3 & 1, 2, 3, 4, 5, 7, 9, 10, 19, 22, 36, 41, 47, 49, 50, 51 & 21, 33, 35, 54 & 29, 37, 39, 58 \\
Dolmino3 & 1, 2, 3, 4, 5, 7, 9, 10, 19, 22, 38, 50, 51, 54 & 6, 21, 33, 35, 36, 49 & 29, 37, 39, 41, 47, 58 \\
SFT3 & 1, 3, 4, 5, 6, 7, 10, 14, 16, 35, 38, 49, 51, 53, 58 & 19, 21, 33, 50, 54 & 2, 9, 37, 41, 47 \\
DPO3 & 1, 2, 3, 4, 5, 6, 7, 8, 9, 10, 14, 16, 19, 21, 22, 29, 33, 35, 36, 37 &  &  \\
Pile & 1, 2, 3, 4, 5, 8, 9, 10, 19, 21, 29, 38, 39, 50, 54 & 7, 22, 35, 36, 49 & 37, 41, 47, 51, 58 \\
RefinedWeb & 1, 2, 3, 4, 5, 6, 7, 8, 9, 10, 14, 16, 19, 21, 22, 29, 33, 35, 36, 37 &  &  \\ \bottomrule
\end{tabular}
\caption{Overlaps in top-20 topics for each stance and dataset. See keywords for topics in Table~\ref{tab:common-topics-keywords}.}
\label{tab:top-topics}
\end{table*}

\begin{table*}[t]
\centering
\footnotesize
\begin{tabular}{@{}ll@{}}
\toprule
Topic 1 & like, time, people, just, new, know, don, make, world, years, right, way \\
Topic 2 & said, police, people, law, news, court, state, new, government, like, year, just \\
Topic 3 & god, people, life, like, just, time, world, don, know, way, think, good \\
Topic 4 & law, people, workers, state, government, work, discrimination, federal, rights, new, act, policy \\
Topic 5 & china, war, government, world, military, states, united, american, countries, people, united states, country \\
Topic 6 & \begin{tabular}[c]{@{}l@{}}sustainable, user, assistant, environmental, climate, request, future, change, climate change, action,\\ planet, o3 request\end{tabular} \\
Topic 7 & \begin{tabular}[c]{@{}l@{}}energy, renewable, solar, gas, climate, carbon, emissions, renewable energy, electricity, fossil, power,\\ wind\end{tabular} \\
Topic 8 & minister, modi, government, bjp, party, said, india, prime minister, prime, leader, narendra, narendra modi \\
Topic 9 & israel, israeli, palestinian, jewish, obama, palestinians, president, state, gaza, said, government, jews \\
Topic 10 & \begin{tabular}[c]{@{}l@{}}trump, president, election, donald, republican, obama, said, donald trump, clinton, campaign, democrats,\\ party\end{tabular} \\
Topic 14 & \begin{tabular}[c]{@{}l@{}}stereotypes, individuals, important, understanding, inclusive, diversity, discrimination, harmful, respect,\\ approach, regardless, diverse\end{tabular} \\
Topic 16 & narrative, world, story, user, global, digital, documentary, request, assistant, film, fictional, tulu\_v3 \\
Topic 19 & city, budget, government, tax, new, public, money, state, said, million, year, plan \\
Topic 21 & gay, lgbtq, people, black, community, white, media, rights, sexual, lgbt, like, women \\
Topic 22 & \begin{tabular}[c]{@{}l@{}}students, education, college, university, universities, student, school, higher education, higher, jobs, job,\\ people\end{tabular} \\
Topic 29 & internet, companies, net neutrality, neutrality, federal, fcc, new, providers, data, service, rules, services \\
Topic 33 & \begin{tabular}[c]{@{}l@{}}energy, income, equity, housing, access, community, renewable, communities, affordable, sustainable,\\ renewable energy, policy\end{tabular} \\
Topic 35 & gender, women, stereotypes, inclusive, education, equality, roles, female, men, support, diverse, diversity \\
Topic 36 & california, food, animals, animal, state, meat, industry, farm, farmers, agriculture, farms, said \\
Topic 37 & market, government, business, companies, prices, economic, price, free, money, profit, tax, businesses \\
Topic 38 & \begin{tabular}[c]{@{}l@{}}request, assistant, user, individuals, harmful, tulu\_v3, stereotypes, understanding, respect, help,\\ respectful, community\end{tabular} \\
Topic 39 & senate, tax, state, pay, house, said, congress, reform, legislation, year, vote, republicans \\
Topic 41 & child, children, law, punishment, criminal, abuse, court, crime, justice, abortion, legal, women \\
Topic 47 & police, gun, officers, guns, firearms, officer, law, law enforcement, enforcement, shooting, firearm, carry \\
Topic 49 & film, films, world, movie, life, water, like, climate, story, people, new, environmental \\
Topic 50 & \begin{tabular}[c]{@{}l@{}}climate, climate change, warming, global, change, global warming, carbon, emissions, scientists,\\ greenhouse, weather, temperatures\end{tabular} \\
Topic 51 & \begin{tabular}[c]{@{}l@{}}education, schools, school, students, educational, children, teachers, parents, cultural, learning, religious,\\ academic\end{tabular} \\
Topic 53 & jane, sam, scene, aaron, just, dialogue, like, int, eyes, let, know, voice \\
Topic 54 & people, media, like, said, black, social, just, social media, racism, women, twitter, police \\
Topic 58 & god, jesus, christ, christian, church, bible, faith, life, christians, jesus christ, lord, love \\ \bottomrule
\end{tabular}
\caption{Topics extracted using Qwen3 Embedding 4B as the embedding model based on a sample of up to 3000 left and right (each) documents across all datasets.}
\label{tab:common-topics-keywords}
\end{table*}

\section{Stance Classification}
\label{appendix:stance-classification}

\subsection{Stance Annotations}
\label{appendix:stance-annotations}

 In order to sample documents which are related to the topics of interest, we first compute centroids of SBERT-encodings (\textsc{all-mpnet-base-v2}) with statements from each policy issue from ProbVAA \citep{ceron-etal-2024-beyond}. We compute the Euclidean distances and based on expert manual inspection, we set a threshold of 0.95 for separating documents belonging to a particular policy issue. We then sample 200 documents from these clusters to be annotated by two annotators. The two annotators are proficient in English. One of them is a master's student in Data Science and Business Analytics and the other is one of the authors of the paper who has extensive experience with political leaning studies. 

Table \ref{tab:annotations-stance} shows the inter-annotator agreement between annotators in the stance annotation task which is considered good for a subjective task \citep{wong2021cross}. The disagreement is solved by the annotator with more extended experience in political science. Table \ref{tab:testset-stance} shows the distribution of the labels.

 The guidelines for the annotations are found in this anonymous link (\url{https://osf.io/caeq3/overview?view_only=036a3331e38d4040a0c4f8e923556b07}). The external annotator was paid 15 euros an hour and completed the task in 22 hours. We will release \textsc{stancepol} under the Open Data Commons Attribution License v1.0 (ODC-By). It is intended for research and educational use. 

\begin{table}[]
    \centering
    \small
    \caption{Distribution of labels in the test set for stance classification with documents from \textsc{Dolma}.}
    \begin{tabular}{lrrr}
    \toprule
    policy & support & oppose & neutral \\
    \midrule
    expanded environ protec. & 21 & 8 & 171 \\
    expanded social welfare & 27 & 15 & 158 \\
    law and order & 14 & 15 & 171 \\
    liberal economic pol. & 40 & 28 & 132 \\
    liberal society & 26 & 16 & 158 \\
    open foreign pol. & 21 & 19 & 160 \\
    restrictive financial pol. & 17 & 27 & 156 \\
    restrictive migration pol. & 11 & 15 & 174 \\
    \bottomrule
    \end{tabular}
    \label{tab:testset-stance}
\end{table}

\subsection{Stance Classification}
\label{appendix:stance-classi}

\subsubsection{Policy Issue Definitions}
\begin{lstlisting}[style=promptstyle]
POLICY_ISSUES = {
  "open_foreign_policy": (
    "Open Foreign Policy",
    "support for international cooperation, "
    "trade, alliances, global institutions"),
  "liberal_economic_policy": (
    "Liberal Economic Policy",
    "free markets, deregulation, privatization, "
    "competition"),
  "restrictive_financial_policy": (
    "Restrictive Financial Policy",
    "budget cuts, fiscal conservatism, debt "
    "reduction, spending restraint"),
  "law_and_order": (
    "Law and Order",
    "policing, criminal justice, public safety, "
    "surveillance"),
  "restrictive_migration_policy": (
    "Restrictive Migration Policy",
    "immigration limits, border control, "
    "deportation"),
  "expanded_environ_protection": (
    "Expanded Environmental Protection",
    "climate action, green energy, regulation "
    "of emissions"),
  "expanded_social_welfare_state": (
    "Expanded Social Welfare State",
    "public healthcare, unemployment insurance, "
    "pensions, welfare benefits"),
  "liberal_society": (
    "Liberal Society",
    "civil rights, minority protections, gender "
    "equality, LGBTQ+ rights"),
}
\end{lstlisting}

\subsubsection{System Prompt}

\begin{lstlisting}[style=promptstyle]
You are an expert political science researcher
specializing in policy stance classification.
\end{lstlisting}

\subsubsection{User Prompt Template}

\begin{lstlisting}[style=promptstyle]
# TASK
Classify the stance expressed in the text toward
the specified policy issue.

TEXT: {text}

POLICY ISSUE: {policy_label}
(i.e., {policy_description})

SUPPORT examples: {support_examples}
OPPOSE examples: {oppose_examples}

INSTRUCTIONS
Analyze the text carefully before answering:

1. Determine whether the text discusses, refers
   to, or clearly implies the specific policy
   issue. A merely related topic does not count.
2. Determine whether the text expresses a
   position toward the issue.
3. If a position is present, decide whether its
   overall thrust supports or opposes the issue.
4. If the opposite stance is equally plausible,
   or if conflicting positions are genuinely
   balanced, classify the stance as "neutral".
5. Use "support" or "oppose" only when the
   stance is explicit or strongly and
   unambiguously implied.
6. Perform this analysis internally. Do not
   include explanations or reasoning in the
   output.

OUTPUT RULES
* If the issue is not discussed, return
  "discussed": false and "stance": "neutral".
* If the issue is discussed but no clear
  position is expressed, return "discussed":
  true and "stance": "neutral".
* Return only valid JSON.

{{ "discussed": true | false,
   "stance": "neutral" | "support" | "oppose" }}
\end{lstlisting}

\begin{table*}[]
    \centering
    \small
    \resizebox{\columnwidth}{!}{%
    \begin{tabular}{lccccc}
    \toprule
    Policy & Maj.Bas. & qwen-P & gpt-P & qwen-R & gpt-R \\
    \midrule
    expanded-environ-prot. & 0.307 & 0.85 & 0.79 & 0.92 & 0.85 \\
    restrictive-migration & 0.310 & 0.84 & 0.86 & 0.96 & 0.99 \\
    restrictive-financial & 0.292 & 0.77 & 0.72 & 0.90 & 0.81 \\
    open-foreign-policy & 0.296 & 0.77 & 0.80 & 0.93 & 0.91 \\
    expanded-social-welfare & 0.294 & 0.77 & 0.77 & 0.89 & 0.86 \\
    liberal-society & 0.294 & 0.75 & 0.83 & 0.79 & 0.74 \\
    liberal-economic-policy & 0.265 & 0.73 & 0.73 & 0.80 & 0.80 \\
    law-and-order & 0.307 & 0.69 & 0.69 & 0.83 & 0.79 \\
    Average & 0.296 & 0.77 & 0.78 & 0.88 & 0.84 \\
    \bottomrule
    \end{tabular}
    }
    \caption{Precision (P) and recall (R) of the stance classification on the test set with Qwen3.6-27B and GPT-5.4-mini, alongside the majority-class baseline F1 macro. Each policy issue has 200 samples in the test set.}
    \label{tab:classification-stance}
\end{table*}

\begin{table*}[]
    \centering
    \small
\begin{tabular}{lrrrrrrrrrrrr}
\toprule
Dataset & \multicolumn{2}{r}{Dolma2} & \multicolumn{2}{r}{Dolmino2} & \multicolumn{2}{r}{Refinedweb} & \multicolumn{2}{r}{Pile} & \multicolumn{2}{r}{Tulu sft 2} & \multicolumn{2}{r}{Dpo mix 2} \\
Stance & sup & opp & sup & opp & sup & opp & sup & opp & sup & opp & sup & opp \\
\midrule
expanded social welfare state & 1427 & 438 & 1181 & 624 & 658 & 310 & 971 & 447 & 181 & 15 & 583 & 71 \\
open foreign policy & 859 & 534 & 1124 & 597 & 475 & 307 & 792 & 559 & 72 & 12 & 373 & 96 \\
expanded environ protection & 1513 & 376 & 2664 & 558 & 839 & 217 & 1061 & 313 & 304 & 9 & 1312 & 73 \\
liberal society & 4219 & 1137 & 3354 & 1055 & 2044 & 641 & 2697 & 887 & 1466 & 43 & 2796 & 221 \\
law and order & 1310 & 1774 & 719 & 1577 & 705 & 919 & 1140 & 1568 & 51 & 115 & 311 & 485 \\
liberal economic policy & 681 & 2134 & 1082 & 2567 & 397 & 1153 & 611 & 1680 & 44 & 108 & 195 & 668 \\
restrictive financial policy & 363 & 1289 & 539 & 1259 & 287 & 689 & 378 & 1046 & 17 & 56 & 50 & 398 \\
restrictive migration policy & 413 & 588 & 293 & 510 & 234 & 280 & 459 & 533 & 17 & 87 & 76 & 160 \\
\bottomrule
\end{tabular}
    \caption{Portion of documents classified as \textit{support} (sup), \textit{oppose} (opp), and \textit{neutral} (neu) by policy issue for each dataset.}
    \label{tab:results-stance-classification-docs}
\end{table*}

\begin{table}[]
    \centering
    \begin{tabular}{llrr}
    \toprule
    Dataset 1 & Dataset 2 & Pearson r & p-value \\
    \midrule
    Refinedweb & Pile & 0.997 & 0.000 \\
    Refinedweb & Dolma3 & 0.956 & 0.000 \\
    Refinedweb & Dolmino3 & 0.938 & 0.001 \\
    Pile & Dolma3 & 0.953 & 0.000 \\
    Pile & Dolmino3 & 0.931 & 0.001 \\
    Dolma3 & Dolmino3 & 0.994 & 0.000 \\
    \hdashline
    Average & & 0.96 & \\
    \bottomrule
    \end{tabular}
    \caption{Pearson correlations between pre-training datasets.}
    \label{tab:pearson-corr-datasets}
\end{table}

\section{Model Behavior}
\label{appendix:model-behavior}

\begin{figure}
    \centering
    \includegraphics[width=\linewidth]{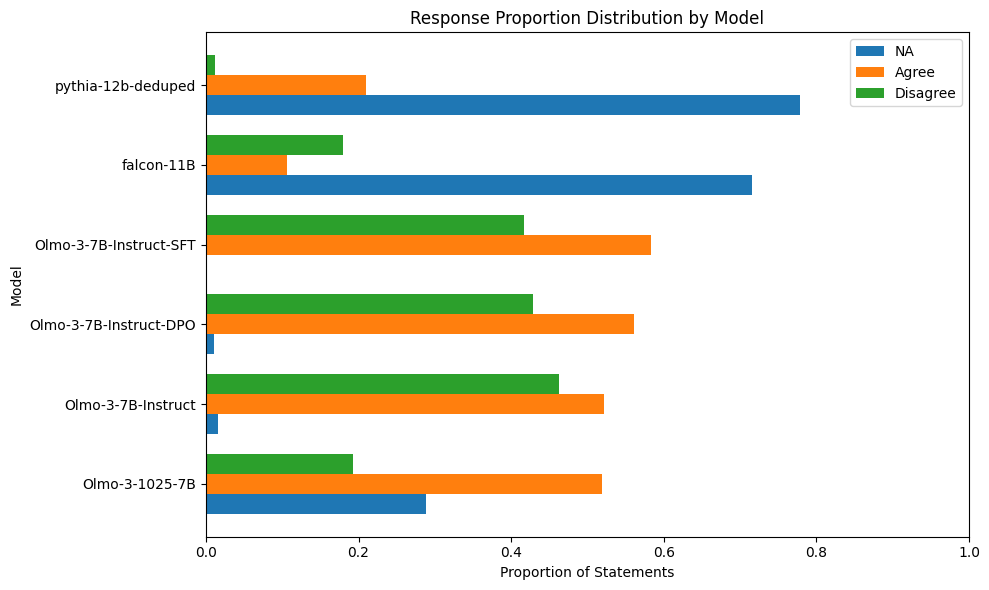}
    \caption{Proportion of answers per model in the model behavior evaluation.}
    \label{fig:num-nas}
\end{figure}

Figure \ref{fig:num-nas} shows the proportion of answers generated when probing models for model behavior. 

\section{Neutral documents}
\label{appendix:neutral-documents}

Topic analysis of neutral documents. To characterize what the neutral portion of each corpus is about, we ran topic modeling over a uniform sample of it. We randomly sampled 5,000 neutral documents from each of six datasets: Dolma 3, Dolmino 3, the OLMo 3 SFT and DPO data, RefinedWeb and the Pile. This gave 30,000 documents pooled into one collection. We fit a Latent Dirichlet Allocation model with 100 topics on bag-of-words counts. The vocabulary was restricted to alphabetic tokens of at least two letters, with English stop words removed, words appearing in more than 50\% or fewer than 5 documents dropped, and a cap of 20,000 terms. We assigned each document to its dominant topic. To name the clusters, we gave GPT-5.4-mini each topic's 15 highest-weight words and its 10 most representative documents. These are the documents assigned to that topic with the highest topic probability, truncated to 800 characters. The model returned a short name (2–5 words). The resulting names, top words and cluster sizes are in Table \ref{tab:neutral-docs}.

\begin{table*}[]
    \centering
    \small
\begin{tabular}{rll}
\toprule
n docs & name & top words \\
\midrule
4436 & Casual Conversation Replies & like, just, don, people, know, good, way, make, time, think \\
1579 & Python List Processing & input, list, python, output, return, code, function, string, example, print \\
1550 & Arithmetic Problem Solving & number, right, left, let, total, given, step, function, problem, equation \\
1411 & Labor and Childbirth Scenes & like, just, eyes, said, hand, head, face, did, let, didn \\
1207 & Cancer Biomarker Studies & patients, cells, study, treatment, disease, activity, type, cancer, studies, associated \\
1055 & Q\&A Responses & answer, com, question, user, request, https, www, yes, http, email \\
964 & Mental Health Support & health, help, care, support, medical, mental, individuals, provide, important \\
937 & Online Service Offers & online, free, features, offer, offers, experience, best, users, services, platform \\
883 & Project Management Work & work, project, management, team, development, services, business, service, job \\
798 & Urban Local Attractions & city, new, local, event, home, park, area, building, street, place \\
700 & Organic Reaction Mechanisms & acid, cell, reaction, protein, carbon, proteins, cells, genes, dna, group \\
674 & Identity and Culture & social, life, world, self, understanding, unique, cultural, role, personal, sense \\
658 & Government and Elections & state, public, government, said, president, states, national, new, news, mr \\
649 & Cloud Database Queries & file, cloud, server, use, using, database, command, query, data, create \\
556 & Contextual Information Requests & information, language, used, context, based, example, document, provide, case \\
548 & Education Institutions & school, students, university, education, student, college, learning, program, class \\
534 & Family Drama & man, life, joan, family, phi, mother, story, father, death, woman \\
501 & Sports Season Recaps & season, big, football, league, just, club, best, hair, night, home \\
435 & Quantum Field Theory & pi, field, space, quantum, gamma, sin, theory, time, epsilon, big \\
431 & Capital Markets & capital, market, million, investment, growth, years, supply, global, demand, billion \\
\bottomrule
\end{tabular}
    \caption{Top 20 clusters for neutral documents.}
    \label{tab:neutral-docs}
\end{table*}

\section{Stances in Documents}

Table \ref{tab:sft-dpo-vs-baseline} shows the two-proportion z-test with Holm correction to test whether each post-training dataset's stance magnitude is significantly higher than the pooled pretraining baseline. Table \ref{tab:pile-refinedweb-vs-dolma-dolmino} shows the results regarding testing whether Pile (P)/RefinedWeb (RW) significantly differ from Dolma3 (D)/Dolmino3 (Dno) in stance magnitude. Table \ref{tab:law-and-order-all-pairs} tests pairwise dataset comparisons for Law and Order. 

\begin{table}[h]
\centering
\small
\begin{tabular}{lcc}
\toprule
 & SFT3 & DPO3 \\
Policy Issue &  &  \\
\midrule
Expanded Social Welfare State & \checkmark & \checkmark \\
Open Foreign Policy & \checkmark & $\times$ \\
Expanded Environmental Protection & \checkmark & $\times$ \\
Liberal Society & \checkmark & \checkmark \\
Law and Order & $\times$ & \checkmark \\
Liberal Economic Policy & $\times$ & $\times$ \\
Restrictive Financial Policy & $\times$ & $\times$ \\
Restrictive Migration Policy & \checkmark & $\times$ \\
\bottomrule
\end{tabular}
\caption{Whether each post-training dataset's stance magnitude is significantly higher than the pooled pretraining baseline (Pile+RefinedWeb+Dolma3+Dolmino), two-proportion z-test with Holm correction, $\alpha=0.05$.}
\label{tab:sft-dpo-vs-baseline}
\end{table}

\begin{table}[h]
\centering
\small
\begin{tabular}{lcccc}
\toprule
Policy Issue & P-D & P-Dno & RW-D & RW-Dno \\
\midrule
Expanded Social W. & \checkmark & $\times$ & \checkmark & $\times$ \\
Open Foreign Policy & \checkmark & \checkmark & \checkmark & \checkmark \\
Expanded Environ. & \checkmark & \checkmark & \checkmark & $\times$ \\
Liberal Society & \checkmark & $\times$ & $\times$ & $\times$ \\
Law and Order & $\times$ & $\times$ & $\times$ & $\times$ \\
Liberal Economic & \checkmark & \checkmark & \checkmark & \checkmark \\
Restrictive Financial & $\times$ & $\times$ & $\times$ & $\times$ \\
Restrictive Migration & $\times$ & \checkmark & $\times$ & $\times$ \\
\bottomrule
\end{tabular}
\caption{Whether Pile (P)/RefinedWeb (RW) significantly differ from Dolma3 (D)/Dolmino3 (Dno) in stance magnitude (two-proportion z-test, Holm-corrected across all pairwise dataset comparisons, $\alpha=0.05$).}
\label{tab:pile-refinedweb-vs-dolma-dolmino}
\end{table}

\begin{table}[h]
\centering
\small
\begin{tabular}{llrc}
\toprule
Dataset 1 & Dataset 2 & $p_{\text{Holm}}$ & Sig. \\
\midrule
Dolma3 & DPO3 & 0.0003 & \checkmark \\
RefinedWeb & DPO3 & 0.0125 & \checkmark \\
SFT3 & DPO3 & 0.0162 & \checkmark \\
Pile & DPO3 & 0.0409 & \checkmark \\
Dolmino & DPO3 & 0.0656 & $\times$ \\
Pile & Dolma3 & 0.1357 & $\times$ \\
Dolma3 & Dolmino & 0.3508 & $\times$ \\
Pile & Dolmino & 1.0000 & $\times$ \\
RefinedWeb & SFT3 & 1.0000 & $\times$ \\
RefinedWeb & Dolmino & 1.0000 & $\times$ \\
RefinedWeb & Dolma3 & 1.0000 & $\times$ \\
Pile & SFT3 & 1.0000 & $\times$ \\
Pile & RefinedWeb & 1.0000 & $\times$ \\
Dolma3 & SFT3 & 1.0000 & $\times$ \\
Dolmino & SFT3 & 1.0000 & $\times$ \\
\bottomrule
\end{tabular}
\caption{All pairwise dataset comparisons for Law and Order (two-proportion z-test, Holm correction scoped to these 15 tests, $\alpha=0.05$).}
\label{tab:law-and-order-all-pairs}
\end{table}

\section{Results with OLMO2}
\label{subsec:olmo2-results}

\begin{figure*}
    \centering
    \begin{subfigure}[b]{0.50\textwidth}
        \includegraphics[width=\linewidth]{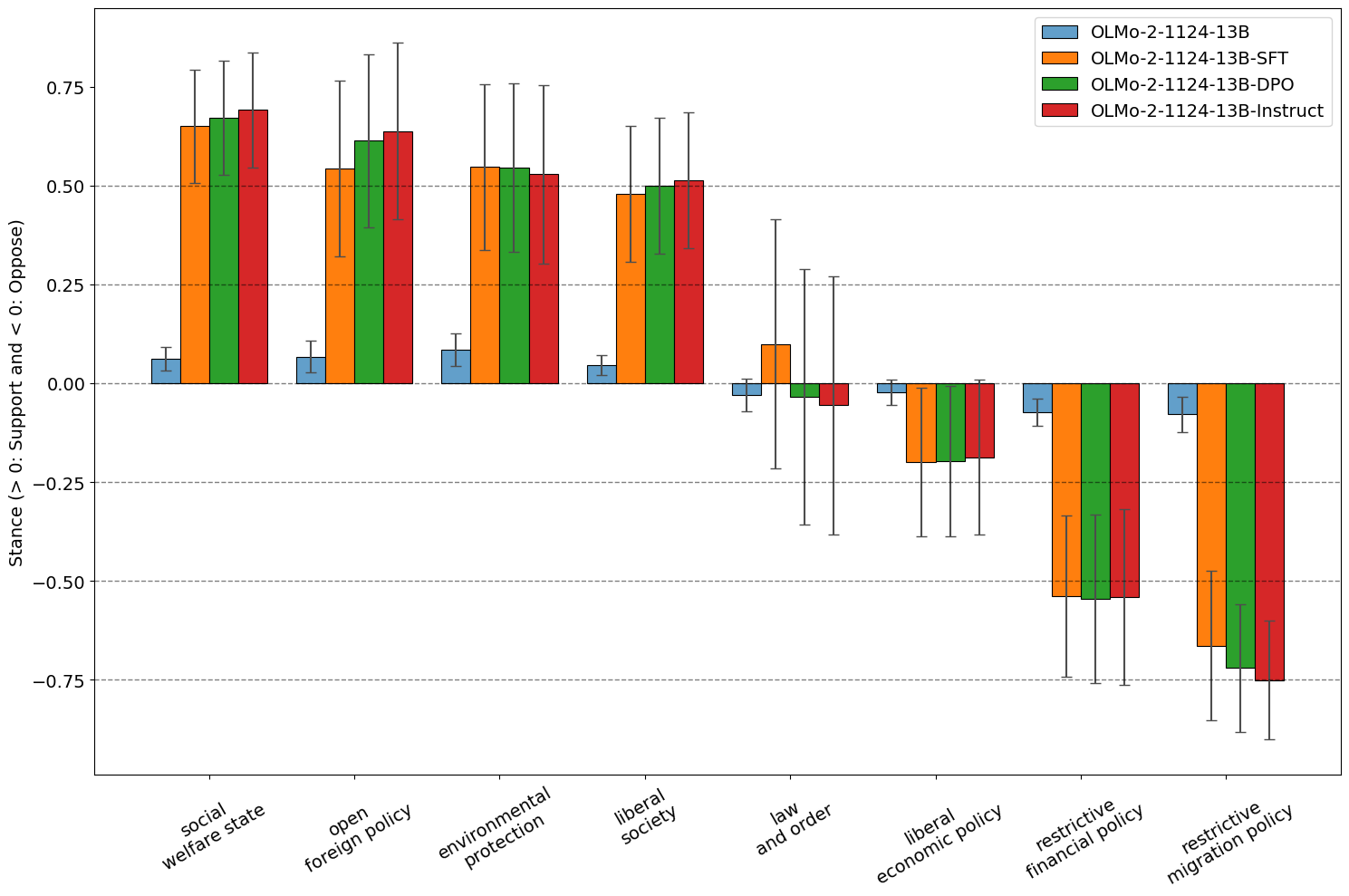}
        \caption{Stance in models}
        \label{fig:olmo-stances}
    \end{subfigure}%
    \hfill
    \begin{subfigure}[b]{0.50\textwidth}
        \includegraphics[width=\linewidth]{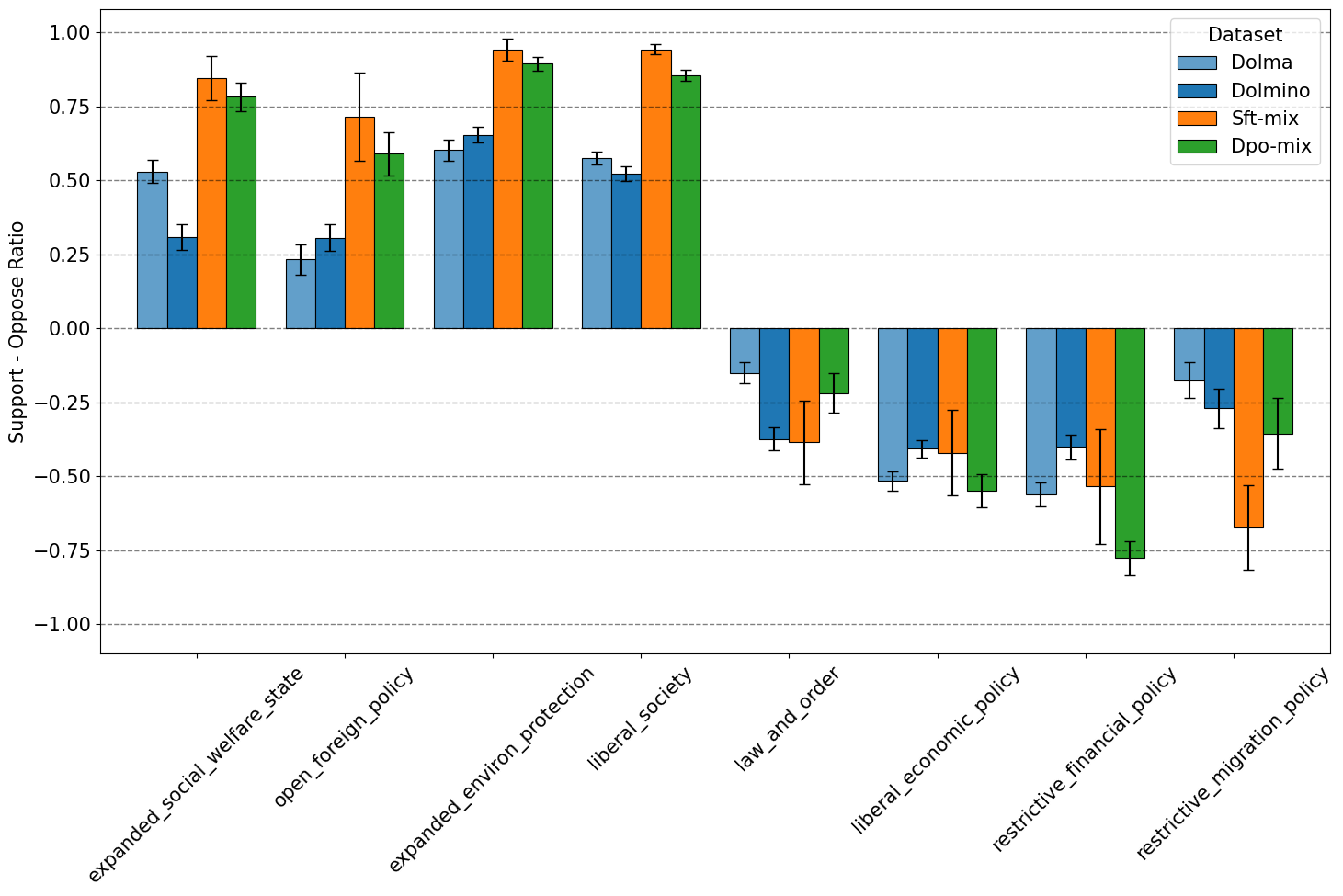}
        \caption{Stances in documents from training data}
        \label{fig:stance-documents}
    \end{subfigure}
    \caption{Stances of the models at different training stages and in the documents from the training data across policy issues. Colors of the bars match models and datasets used to train them at that stage. \textsc{Dolma} and \textsc{Dolmino} are in blue because they are both used in the base model. Error bars show the 95\% confidence intervals for the stance scores.}
    \label{fig:stance-documents-model2}
\end{figure*}

Tables \ref{tab:olmo2-datasets} and Figure \ref{fig:stance-documents-model2} show the results of \textsc{OLMO2} which follow very similar patterns from OLMO3.

\begin{table*}
\caption{Training datasets from OLMO2 with the number of sampled documents and their proportion classified as left- or right-leaning. The remaining documents in the classification results were classified as \textit{neutral}.}
\label{tab:olmo2-datasets}
\centering
\resizebox{\textwidth}{!}{%
\begin{tabular}{cccc c cc}
\cmidrule(lr){1-4} \cmidrule(lr){6-7}
\multicolumn{4}{c}{{\textbf{Documents for analysis}}} & & \multicolumn{2}{c}{{\textbf{Classification results}}} \\
\cmidrule(lr){1-4} \cmidrule(lr){6-7}
\textbf{Dataset name} & \textbf{Total size} & \textbf{Training and model} & \textbf{\# Docs} & & \textbf{\# (\%) Left} & \textbf{\# (\%) Right} \\
\cmidrule(lr){1-4} \cmidrule(lr){6-7}
\makecell{ \textsc{\textbf{Dolma}} \\ \textsc{olmo-mix-1124}} & 22.4TB & \makecell{Pre-training \\ \textsc{OLMo-2-1124-13B}}  & 299,915 & & 12,790 (4.2\%) & 4,644 (1.5\%) \\
\makecell{\textsc{\textbf{Dolmino}} \\ \textsc{dolmino-mix-1124}} & 5.14TB & \makecell{Mid-pre-training \\ \textsc{OLMo-2-1124-13B}} & 200,000 & & 10,537 (5.2\%) & 3,374 (1.6\%) \\
\makecell{\textsc{\textbf{SFT-Mix}}  \\ \textsc{tulu-3-sft-olmo-2-mixture} } & 1.41GB & \makecell{Supervised finetuning (SFT) \\ \textsc{OLMo-2-1124-13B-SFT}} & 193,447 & & 2,863 (1.5\%) & 242 (0.12\%) \\
\makecell{\textsc{\textbf{DPO-Mix}} \\ \textsc{olmo-2-1124-13b-preference-mix} } & 1.46GB & \makecell{Direct Preference Optim. (DPO) \\ \textsc{OLMo-2-1124-13B-DPO}} & 313,609 & & 5,792 (1.8\%) & 920 (0.29\%) \\
\cmidrule(lr){1-4} \cmidrule(lr){6-7}
\end{tabular}
}
\end{table*}

\section{LLM usage}

We have used Claude Code with Opus 5 to aid in the development of the code base for the project and ChatGPT to paraphrase sentences in the main document. 

\end{document}